\documentclass[lettersize,journal]{IEEEtran}
\usepackage{algorithmic}
\usepackage{algorithm}
\usepackage{array}
\usepackage{amsmath,amsfonts}
\usepackage{amsmath,bm}
\usepackage{amsthm}
\usepackage{amssymb}
\usepackage{booktabs}
\usepackage{bm}
\usepackage{color}
\usepackage{cite}
\usepackage{caption}
\usepackage{flushend}
\usepackage{float}
\usepackage[colorlinks,bookmarksopen,bookmarksnumbered,citecolor=green,linkcolor=red,urlcolor=green]{hyperref}
\usepackage{graphicx}
\usepackage{mathrsfs}
\usepackage{mathtools}
\usepackage{multirow}
\usepackage{stfloats}
\usepackage[caption=false,font=normalsize,labelfont=sf,textfont=sf]{subfig}
\usepackage{times}
\usepackage{textcomp}
\usepackage{threeparttable}
\usepackage{url}
\usepackage{verbatim}
\usepackage[table,xcdraw]{xcolor}

\def\BibTeX{{\rm B\kern-.05em{\sc i\kern-.025em b}\kern-.08em
    T\kern-.1667em\lower.7ex\hbox{E}\kern-.125emX}}

\begin{document}
\title{Implicit Motion-Compensated Network for Unsupervised Video Object Segmentation}
\author{Lin~Xi,
	Weihai~Chen*,
	Xingming~Wu,
	Zhong~Liu,
	and Zhengguo~Li
	\thanks{L. Xi, W. Chen, X. Wu, and Z. Liu are with the School of Automation Science and Electrical Engineering, Beihang University, Beijing 100191, China (e-mail: xilin1991@buaa.edu.cn; whchen@buaa.edu.cn; wxmbuaa@163.com; liuzhong@buaa.edu.cn).}
	\thanks{Z. Li is with the SRO Department, Institute for Infocomm Research, Agency for Science, Technology and Research (A*STAR), Singapore 138632 (e-mail: ezgli@i2r.a-star.edu.sg).}
	\thanks{*Corresponding author: Weihai Chen.}}

\maketitle

\begin{abstract}
	Unsupervised video object segmentation (UVOS) aims at automatically separating the primary foreground object(s) from the background in a video sequence. Existing UVOS methods either lack robustness when there are visually similar surroundings (appearance-based) or suffer from deterioration in the quality of their predictions because of dynamic background and inaccurate flow (flow-based). To overcome the limitations, we propose an implicit motion-compensated network (IMCNet) combining complementary cues (\emph{i.e.}, appearance and motion) with aligned motion information from the adjacent frames to the current frame at the feature level without estimating optical flows. The proposed IMCNet consists of an affinity computing module (ACM), an attention propagation module (APM), and a motion compensation module (MCM). The light-weight ACM extracts commonality between neighboring input frames based on appearance features. The APM then transmits global correlation in a top-down manner. Through coarse-to-fine iterative inspiring, the APM will refine object regions from multiple resolutions so as to efficiently avoid losing details. Finally, the MCM aligns motion information from temporally adjacent frames to the current frame which achieves implicit motion compensation at the feature level. We perform extensive experiments on $\textit{DAVIS}_{\textit{16}}$ and $\textit{YouTube-Objects}$. Our network achieves favorable performance while running at a faster speed compared to the state-of-the-art methods. Our code is available at \href{https://github.com/xilin1991/IMCNet}{https://github.com/xilin1991/IMCNet}.
\end{abstract}


\section{Introduction}\label{sec:intro}

\begin{figure}[!h]
	\begin{center}
		\includegraphics[width=1.0\linewidth]{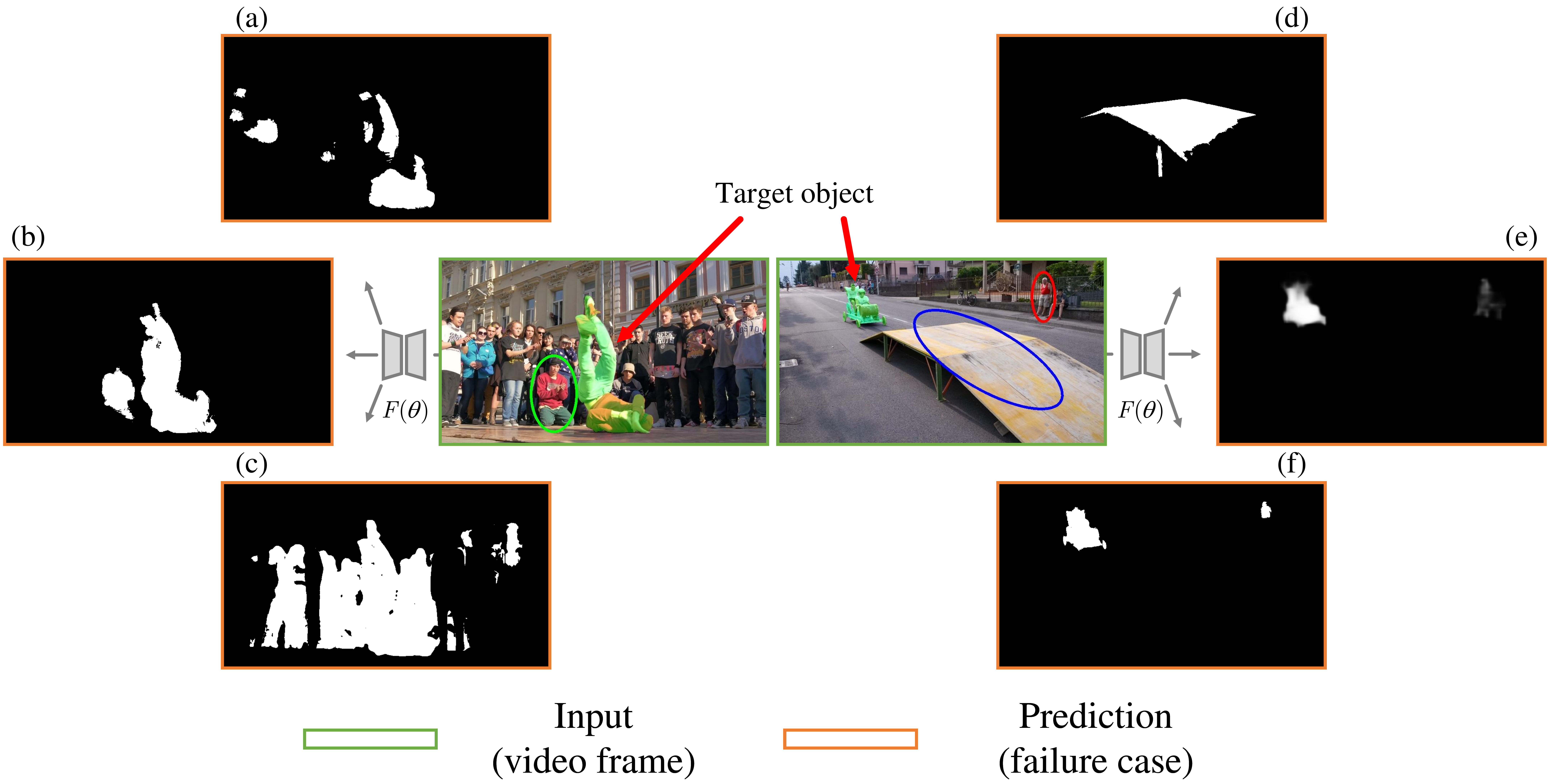}
	\end{center}
	\caption{Illustration of the failure case of existing UVOS methods. The images with a green border as the input frames fed into the models $F(\theta)$ to produce object masks inside the orange border. The objects in the green, blue and red ellipse box are respectively similar surroundings, misleading objects, and motion artifacts. Note that target objects (ground-truth) are highlighted by green translucent masks.}
	\label{fig:Head}
\end{figure}

\IEEEPARstart{T}{he} goal of unsupervised video object segmentation (UVOS) is to identify and segment the most visually prominent object(s) from the background in videos. The UVOS has attracted significant attention owing to its widespread applications in content-based video retrieval \cite{VidRet}, video matting \cite{VidMat}, video editing \cite{VidEdt}, video analysis \cite{VidAnalsis}, and video compression \cite{VidCom1, VidCom2, VidCom3, VidCom4, VidCom5, VidCom6}. Different from the semi-supervised \cite{OSVOS, MaskTrack, RVOS, ZIYANG} and weakly-supervised \cite{Weakly1, Weakly2} video object segmentation, which has a full mask of the object(s) of interest in the first frame of a video sequence, the UVOS is more challenging because of the lack of user interactions. 

Primary object candidates that span through the whole video sequence are provided by the algorithm on the UVOS. This primary foreground object(s) should capture human attention when the whole video sequence is watched, \emph{i.e.}, objects that are more likely to be followed by the human gaze. Inspired by a biological mechanism known as human attention \cite{HumanAtt}, the UVOS system should have remarkable motion perception capabilities to quickly orient gaze to moving objects in dynamic scenes. We argue that the primary object(s) in a video should be (i) the most distinguishable in a single frame, (ii) repeatedly appearing throughout the video sequence, and (iii) moving objects in the video. The former represents the intra-frame discriminability (local dependence), and the latter two are inter-frame consistency (global dependence). Fully exploiting these three properties is crucial for determining the primary video object(s).

For example, methods in \cite{SOD1, SOD2, SOD3} can work well for discovering saliency objects in static images. However, they may not be applicable for UVOS tasks due to motions, occlusion, and distractor objects, as shown in Fig. \ref{fig:Head} (d). Therefore, it is necessary to make use of inter-frame consistency (global dependence) to make up for the drawbacks caused by ignoring temporal correlation. Earlier traditional methods \cite{Sal1, MoBo2, Traj6} have been investigated from these perspectives, it is inadequately considered by current deep learning-based solutions. The deep learning-based methods \cite{FSEG, LVO, SFL} and reinforcement learning-based methods \cite{RL1, RL2} boost the performance of UVOS systems, and yet still face challenges considering local variations and global consistency (uniformity), meanwhile. These works mainly proceed in two directions. One line \cite{LVO, LSMO, FSEG, MotAdapt, MATNet, EpONet, FEMNet, DyStaB} is with a learned model of multi-modality inputs. They typically used optical flows as an additional modality to capture motion cues, which leverage the global temporal motion information cross frames. In general, these multi-modality-based methods introduce an additional pre-processing stage to predict the optical flow. However, due to the accuracy of optical flow prediction, the quality of the UVOS model could deteriorate over time. The noise introduced by the optical flow may misguide the UVOS models to predict the primary object(s) incorrectly as shown in Fig. \ref{fig:Head} (a), (f). Another limitation of those methods is that they ignore local discriminability, such as texture features. The other direction is based on single-modality \cite{SFL, LMP, PDB, COSNet1, COSNet2, ADNet, AGNN, AGS, WCSNet, DFNet}, which matches the appearance of inter-frame to exploit long-term correlations from a global perspective. However, they incur significant computational costs on the condition that non-local operation conducts matrix multiplication operation. More importantly, due to only mining the matched pixels across the frames, previous appearance-based methods may be wrongly taken visually similar surroundings as the primary object(s) (Fig. \ref{fig:Head} (b), (c), (e)). 

In this paper, we propose an \textit{Implicit Motion-Compensated Network} (IMCNet) for the UVOS. The IMCNet takes several frames from the same video as inputs to mine the long-term correlations and align temporally adjacent frames to the current frame for implicit motion compensation at the feature level. The proposed model consists of an affinity computing module (ACM), an attention propagation module (APM), and a motion compensation module (MCM). In our method, the two main objectives are achieved, mining local dependence and offsetting global dependence from adjacent frames. For mining local dependence, the light-weight ACM is designed, it is extended from a \textit{co-attention} \cite{CoAtt1, CoAtt2} mechanism. Instead of computing affinity in the feature space, a novel key-value embedding module is used in ACM to largely reduce the computational cost and meet high performance. \textit{Through this process, we will see in the experimental results that the IMCNet is not only more robust but also more efficient than the algorithm in} \cite{COSNet1, COSNet2, ADNet, AGNN, AGS, WCSNet, DFNet}. Then, the APM is applied to retain the primary object(s) at each level of the top-down decoder architecture. Through coarse-to-fine iterative inspiring, the APM will refine object regions from multiple resolutions so as to efficiently avoid losing details. Finally, the MCM is proposed to implicitly capture global dependence from adjacent frames. Unlike the previous optical flow-based UVOS methods \cite{MATNet, FEMNet, DyStaB}, our approach can adaptively offset the current moment from neighboring frames at the feature level without explicit motion estimation. \textit{This is another advantage of our IMCNet}.

Inspired by the deformable convolution \cite{DCN, DCNV2}, our MCM uses features from the adjacent frames to dynamically predict offsets of sampling convolution kernels. These dynamic kernels are gradually applied to neighboring features, and offsets are propagated to the current frame to facilitate precise motion compensation, similar to the motion transition based on optical flow in multi-modality models \cite{MATNet, FEMNet, DyStaB, TransportNet,  RTNet}. Moreover, an additional temporal-spatial fusion is employed after the cascading deformable operation to further improve the robustness of compensation. A temporal attention by computing the similarity between the feature of compensation and the current frame is applied to utilize motion cues better. After fusing with temporal attention, a spatial attention operation is further introduced to automatically select relatively important features effectively in spatial channels. In addition, a joint training strategy is introduced for training our IMCNet to effectively learn the discriminative features of the primary object(s). It helps our model focus more on pixels of the primary object(s) with the guidance of the saliency information, thus filtering out the similar surrounding noise.

Our contributions in this paper are several folds:
\begin{itemize}
	\item A novel \textit{Implicit Motion-Compensated Network} (IMCNet) combining complementary cues (\emph{i.e.}, appearance and motion) is proposed to overcome the existing method's drawback. This is achieved by a light-weight ACM, APM, and MCM. Extensive experiments on $\textit{DAVIS}_{\textit{16}}$ \cite{DAVIS2016} and $\textit{YouTube-Objects}$ \cite{YouTubeObj} show IMCNet achieves favorable performance while running at faster speed and using much fewer parameters compared to the state-of-the-arts.
	\item A light-weight ACM which embeds a \textit{co-attention} mechanism with a novel key-value component is introduced to explore the consistent representation from a global perspective, while at the same time helping to capture appearance features within inter-frames.
	\item A novel MCM is incorporated for modeling motion information of salient objects within multiple input frames, leading to a more powerful moving object pattern recognition framework.
	\item A new joint training strategy is proposed to train our models on UVOS and SOD datasets, to enhance the representation ability of local discriminability.
\end{itemize}

\section{Proposed method}\label{sec:met}

\begin{figure*}[htbp]
	\begin{center}
		\includegraphics[width=0.90\textwidth]{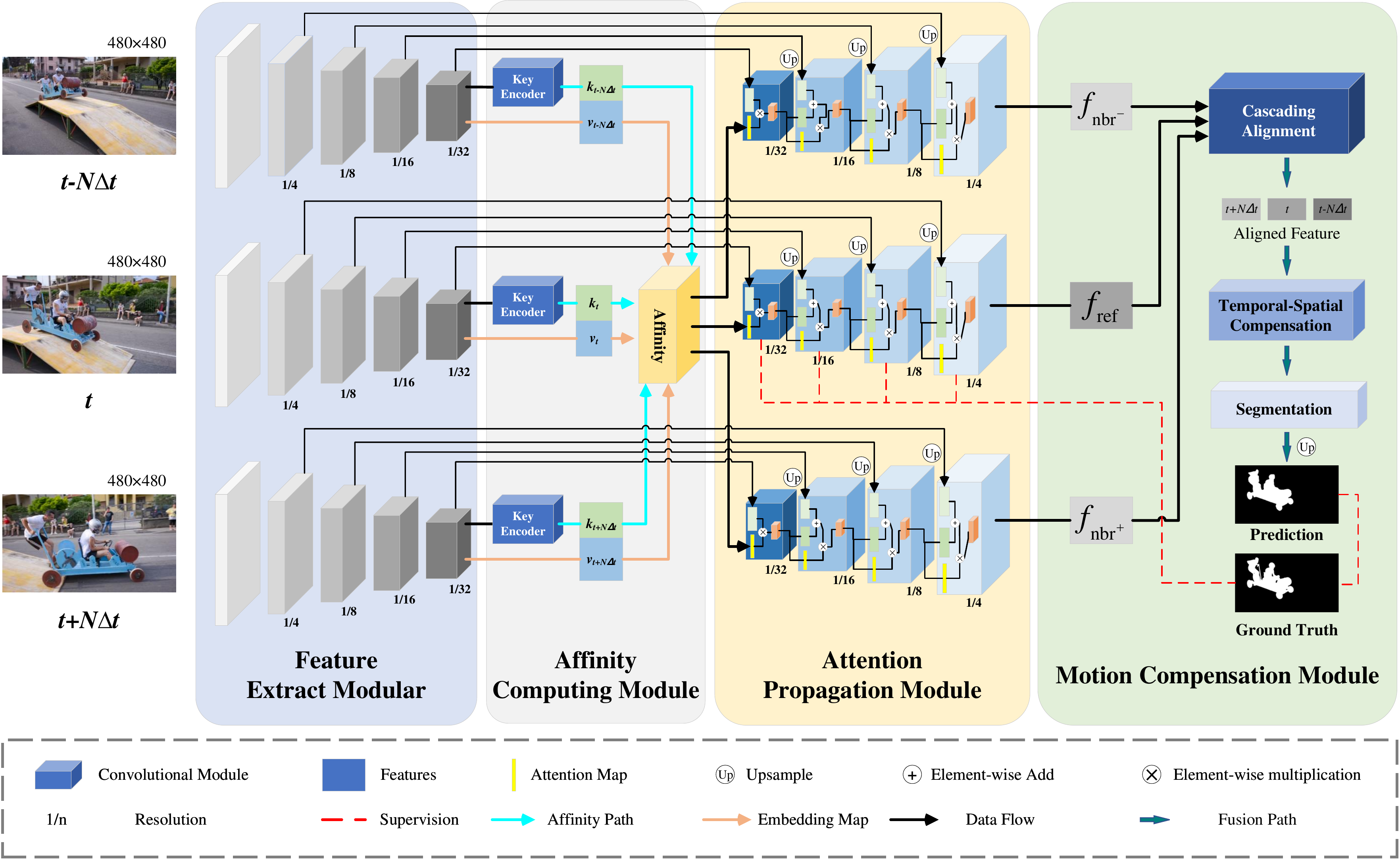}
	\end{center}
	\caption{The overview of the proposed IMCNet. The $2N+1$ consecutive frames $\{I_{i}\}_{i=t-N\Delta t}^{t+N\Delta t}$ are first fed into the FEM to extract embedding of each frame as $V_{i}^{2} \sim V_{i}^{5}$. Then, the ACM computes the affinity of each feature $V_{i}^{5}$ that summarizes the global dependence among input frames $I_{i}$. The attention enhanced features $Z_{i}$ are further fed into the APM to transmit global dependence via a top-down manner. Finally, the output of the top-down decoder is passed to the MCM to facilitate obtaining the final segmentation result $\hat{M}_{t}$. Here, the three key encoders for the center and neighboring frames are parameter-shared.}
	\label{fig:flowchart}
\end{figure*}

An end-to-end deep neural network, \emph{i.e.}, IMCNet, is proposed to mine the long-term correlations and implicitly encodes motion cues in feature level across multi-frames for the UVOS. The overall framework of the proposed IMCNet is shown in Fig. \ref{fig:flowchart}. The IMCNet is designed as an encoder-decoder fashion since this kind of architecture is able to preserve low-level details to refine high-level global contexts. The proposed model comprises four key modules: feature extract modular (FEM) (\S \ref{subsec:31}), ACM (\S \ref{subsec:32}), APM (\S \ref{subsec:33}), and MCM (\S \ref{subsec:34}).  Given $2N+1$ consecutive frames $\{I_{i} \in \mathbb{R}^{w \times h \times 3}\}_{i=t-N\Delta t}^{t+N\Delta t}$ ($N\in\mathbb{N}^{*}$) with an interval of $\Delta t$ frames, the middle frame $I_{t}$ is selected as the center frame and the other frames are neighboring frames, and $w \times h$ indicates the spatial resolution of images. The aim of our model $\mathcal{F}_{\theta}$ is to estimate the corresponding objects masks $\hat{M}_{t} \in \{0, 1\}^{w \times h}$.

\subsection{Feature extract modular}\label{subsec:31}

Our FEM relies on a parallel structure encoder to jointly embed adjacent frames, which has been proven effective in many time-related video tasks. The shared feature encoder, which is pre-trained using the ImageNet \cite{ImageNet} and DUTS \cite{DUTS}, takes several frames as inputs. We take the last four convolutional blocks of the ResNet \cite{ResNet} as the backbone for each stream. The FEM $E_{\theta}$ takes consecutive frames $\{I_{i} \in \mathbb{R}^{w \times h \times 3}\}_{i=t-N\Delta t}^{t+N\Delta t}$ as inputs, including the center frame $I_{t}$ and neighboring frames, for exploring a consensus representation in a high-dimensional space. We denote the extracted embeddings of $I_{i}$ as $V_{i}^{l} \in \mathbb{R}^{\frac{w}{s} \times \frac{h}{s} \times C_{v}^{l}}$, where $l$ indicates the $l$-th ($l \in \{2, 3, 4, 5\}$) residual stage, $s \in \{4, 8, 16, 32\}$ is scales, and $C_{v}^{l}$ represents the feature map channels, respectively.

\subsection{Affinity computing module} \label{subsec:32}

To focus more on the primary object(s) in $I_{t}$, we delve into the intra-frame features supported by neighboring frames. In detail, we build a key encoder module to extract independently a key feature for each frame and is symmetric in the center and neighboring frames\footnote{Key encoder processing on features from FEM does not depend on whether they come from center or neighboring frames. The same key encoder encodes the center and neighboring frames into key maps instead of several independent key encoders.}. The rationales are that 1) Global dependence is computationally efficient to extract from key features than the value ($V_{i}^{5}$), and 2) Global dependence exists in key features without redundancy, and there is a lot of distraction on value features. The ACM then takes two-stream data (\emph{i.e.} key and value features) of each moment as inputs to compute similarity.

\textbf{Key encoder}. We take $V_{i}^{5}$ features from the FEM as inputs of the key encoder. Both the center and the neighboring frames are first encoded into key maps $\{K_{i} \in \mathbb{R}^{{w}' \times {h}' \times C_{k}}\}_{i=t-N\Delta t}^{t+N\Delta t}$ through the residual block, where ${w}'$ and ${h}'$ is $1/32$ smaller than the input image ($w \times h$) and $C_{k}$ is set as 64. It can be formally represented as:
\begin{equation}\label{eq:key:1}
	K_{i}=\xi_{\theta}(V_{i}^{5}),
\end{equation}
where $\xi_{\theta}$ is typically composed of two $3\times3$ convolutional layers followed by a ReLU activation as a projection head from the backbone feature to the key space ($C_{k}$ dimensional).

\begin{figure}[htbp]
	\begin{center}
		\includegraphics[width=0.9\linewidth]{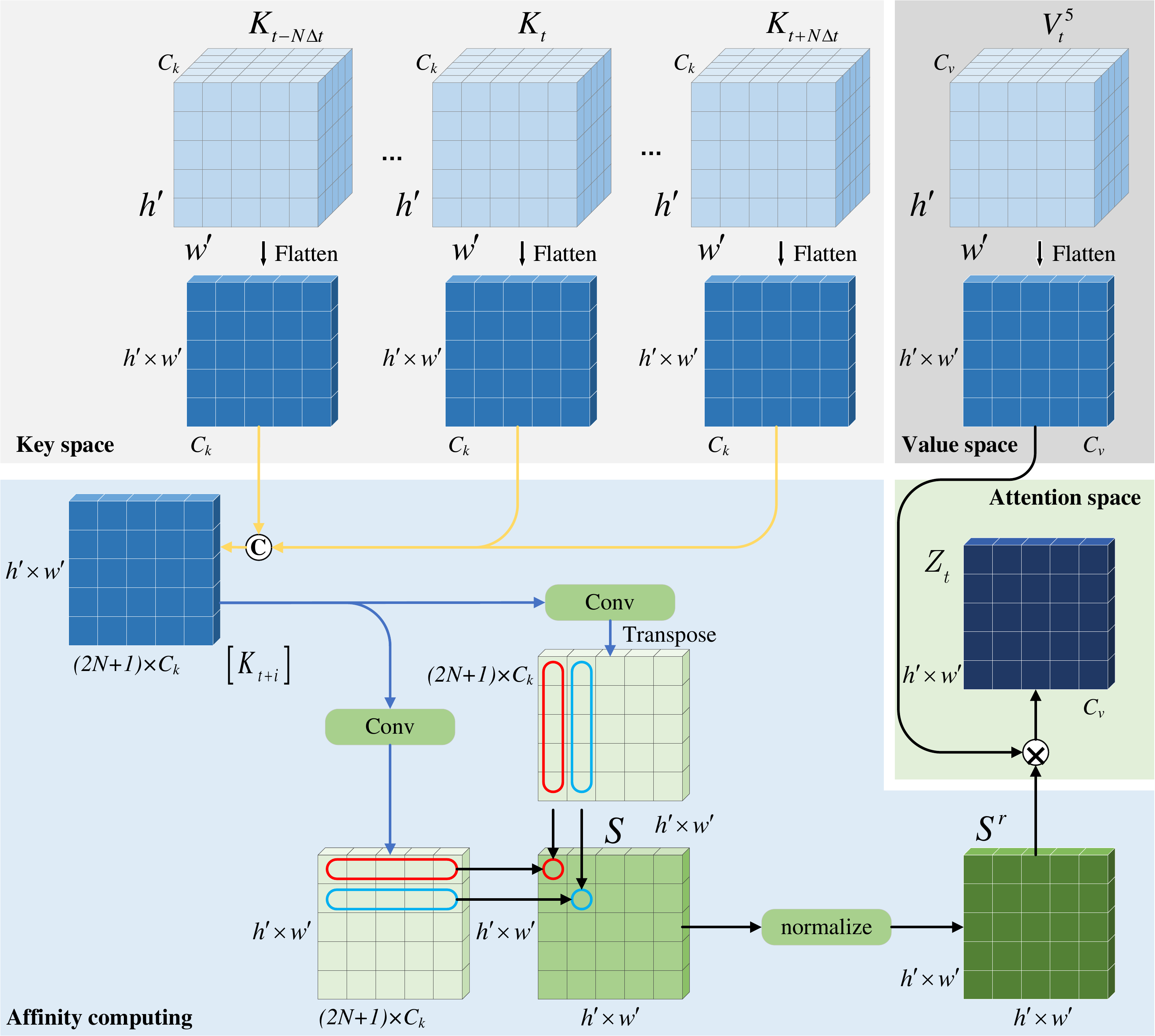}
	\end{center}
	\caption{Illustration of affinity computing process. Taking center moment $t$ as an example, the key features $K_{i}$ are flattened into matrices and used to compute the affinity matrix $S$ via Eq. \ref{eq:ac:2}. Then, the affinity matrix $S$ is normalized via Eq. \ref{eq:ac:3} as attention weight (\emph{i.e.}, $S^{r}$). Finally, the value features $V_{t}^{5}$ are post-multiplied by the $S^{r}$ to compute the attention enhanced feature $Z_{t}$ (Eq. \ref{eq:ac:4}).}
	\label{fig:affcom}
\end{figure}

\textbf{Affinity computing}. Similarities among key features of the center and neighboring frames are computed to determine when-and-where to retrieve relevant primary object(s) correlation from input frames. Key features $K_{i}$ are learned to represent the correlations among the center and neighboring frames in their feature embedding space. Value features $\{V_{i}^{5}\in\mathbb{R}^{{w}'\times{h}'\times{C_{v}}}\}_{i=t-N\Delta t}^{t+N\Delta t}$ store detailed information for producing the object mask prediction. As shown in Fig. \ref{fig:affcom}, given consecutive embedding features $K_{i}$ from the same video, we first compute the affinity matrix $S \in \mathbb{R}^{{w}' \times {h}'}$ among input frames:
\begin{equation}\label{eq:ac:1}
	S=\left [K_{i}\right ]^{\top} \cdot W \cdot \left [K_{i}\right ], i \in \left [t-N\Delta t:t+N\Delta t\right ],
\end{equation}
where [$\cdot$] denotes the concatenation operation and $W \in \mathbb{R}^{C_{k} \times C_{k}}$ is a trainable weight matrix. The affinity matrix $S$ can effectively capture global dependence between the feature space of input images. In order to keep consistency among the video frames, $W$ is approximately factorized into two invertible matrices $P \in \mathbb{R}^{C_{k} \times C_{k}}$ and $Q \in \mathbb{R}^{C_{k} \times C_{k}}$. Then, Eq. \ref{eq:ac:1} can be rewritten as:
\begin{equation}\label{eq:ac:2}
	\begin{aligned}
		S&=\left [K_{i}\right ]^{\top}\cdot P \cdot Q^{\top} \cdot \left [K_{i}\right ]\\
		&=\left( P^{\top}\cdot\left [K_{i}\right ] \right)^{\top}\cdot\left(Q^{\top} \cdot\left[K_{i}\right]\right),
	\end{aligned}
\end{equation}
where $i \in \left [t-N\Delta t:t+N\Delta t\right ]$. This operation is equivalent to applying channel-wise feature transformations to key features $K_{i}$ before computing the similarity.

After obtaining the affinity matrix $S$, as shown in the light blue area in Fig. \ref{fig:affcom}, we normalize $S$ row-wise with a softmax function to derive an attention map $S^{r}$ conditioned on neighboring images and achieve enhanced features.
\begin{equation}\label{eq:ac:3}
	S^{r}=\frac{\mathrm{exp}(S_{ij})}{\sum_{n}(\mathrm{exp}(S_{nj}))}\in\left[0, 1\right]^{({w}'{h}')\times({w}'{h}')}.
\end{equation}
With the normalized affinity matrix $S^{r}$, the attention summaries for the feature embedding $Z_{i}$ for each frame can be computed as a weighted sum of the value feature $V_{i}^{5}$ with efficient matrix multiplication (see the green areas in Fig. \ref{fig:affcom}):
\begin{equation}\label{eq:ac:4}
	Z_{i}=V_{i}^{5} \cdot S^{r}\in\mathbb{R}^{{w}'\times{h}'\times C_{v}},
\end{equation}
which is then passed to the APM.

\subsection{Attention propagation module}\label{subsec:33}

\begin{figure}[htbp]
	\begin{center}
		\includegraphics[width=0.9\linewidth]{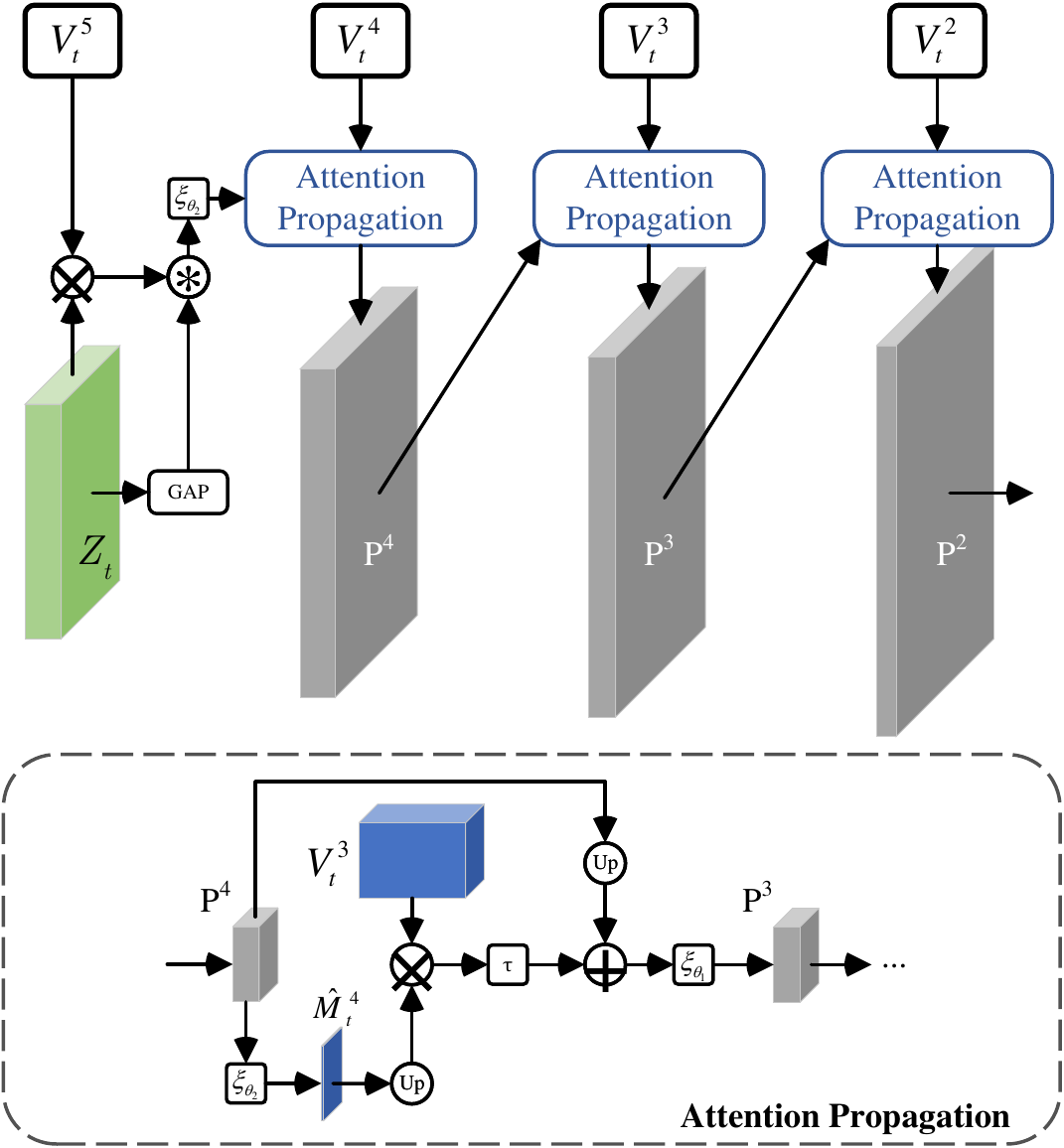}
	\end{center}
	\caption{The proposed APM. Attention propagation operation is implemented by Eq. \ref{eq:apm:1}. The computational graph of center moment $t$ is taken as an example. Here, `$\otimes$', `$\circledast$', `$\oplus$', and `$\mathrm{Up}$' indicate element-wise multiplication, element-wise multiplication with broadcasting, element-wise addition, and bilinear upsampling, respectively. The `$\mathrm{GAP}$' denotes the global average pooling operation.}
	\label{fig:apmodule}
\end{figure}

In general, the FPN \cite{FPN} decoder is built upon the bottom-up backbone, and the induced high-level features will be gradually diluted when transmitted to lower layers. To this end, we propose an APM to connect encoder-decoder pairs of our IMCNet, as shown in Fig. \ref{fig:apmodule}. The APM progressively refines the skip-connection of FEM via higher-level prediction and can locate primary object(s) more accurately without the interference of irrelevant surroundings. Specifically, for each layer, the skip-connection of FEM $V_{i}^{l}$ is guided by the higher-level prediction $\hat{M}_{i}^{l+1}$ to produce the final feature $P_{i}^{l}$. At the top of the APM, we take $V_{i}^{5}$ weighted by the enhanced feature $Z_{i}$ as an immediate feature $P_{i}^{5}$ to predict the first guide map $\hat{M}_{i}^{5}$:
\begin{equation}\label{eq:apm:head}
	\left\{\begin{matrix}
		\begin{aligned}
			P_{i}^{5}&= (V_{i}^{5}\ast Z_{i}) \circledast \mathrm{GAP}(Z_{i}) \\
			\hat{M_{i}^{5}}&=\sigma(\xi_{\theta_{2}}^{5}(P_{i}^{5})),
		\end{aligned}
	\end{matrix}\right.
\end{equation}
where $\ast$, $\circledast$, and $\mathrm{GAP}(\cdot)$ are element-wise multiplication, element-wise
multiplication with broadcasting, and global average pooling operation. $\xi_{\theta_{2}}^{5}(\cdot)$ is implemented by two convolutional layers followed by a sigmoid layer. After that, the APM further fed the guide map into the next layer in a recursive manner:
\begin{equation}\label{eq:apm:1}
	\left\{\begin{matrix}
		\begin{aligned}
			\tilde{V}_{i}^{l}&=(\mathrm{up}(\hat{M_{i}}^{l+1})\ast V_{i}^{l})\\ 
			P_{i}^{l}&=\xi_{\theta_{1}}^{l}(\mathrm{up}(P_{i}^{l+1})+\tau_{\theta}^{l}(\tilde{V}_{i}^{l})), l\in\left\{4,3,2\right\},\\ 
			\hat{M_{i}^{l}}&=\sigma(\xi_{\theta_{2}}^{l}(P_{i}^{l}))
		\end{aligned}
	\end{matrix}\right.
\end{equation}
where `$\ast$' is the element-wise multiplication. $\mathrm{up}(\cdot)$ is the up-sampling operation with stride 2 via bilinear interpolation. $\xi_{\theta_{1}}(\cdot)$ is typically composed of a residual block, with 64 kernels. It is similar to lateral connections of FPN \cite{FPN}. The feature maps $\tilde{V}_{i}^{l}$ are then enhanced with features from the bottom-up pathway via lateral connection, and then $\xi_{\theta_{1}}(\cdot)$ refine it to obtain $P_{i}^{l}$. $\tau(\cdot)$ consists of a head convolutional layer and a residual block and reduces the refined features $\tilde{V}_{i}^{l}$ to 64 channels. As the last step in the APM, we adopt two convolutional layers $\xi_{\theta_{2}}(\cdot)$ followed by a sigmoid layer $\sigma$ to predict the side masks $\hat{M}_{i}^{l}$ for deep supervision. The side masks $\hat{M}_{i}^{l}$ can effectively hold the attention on the primary object(s) in each layer, as presented in the ablation study of section \ref{sec:exp}. The last $P_{i}^{2}$ is the final output into the MCM for learning motion information. $P_{t}^{2}$, named as $f_{\mathrm{ref}}$, represents the output of the APM, in which input center frame $I_{t}$. $f_{\mathrm{nbr}^{\left\{+, -\right\}}}$ denote $\left\{P_{i}^{2}\right\}_{i=t+\Delta t}^{t+N\Delta t}$ and $\left\{P_{i}^{2}\right\}_{i=t-N\Delta t}^{t-\Delta t}$, respectively (\emph{i.e.}, $+$ is the moment after the center frame, and vice versa).

\subsection{Motion compensation module}\label{subsec:34}

The MCM is proposed to align the features from neighboring frames to the center frame's feature space for implicit motion compensation. As illustrated in Fig. \ref{fig:flowchart}, our MCM includes three sub-modules: a cascading alignment, a temporal-spatial compensation, and a segmentation.

\begin{figure}[htbp]
	\begin{center}
		\includegraphics[width=0.9\linewidth]{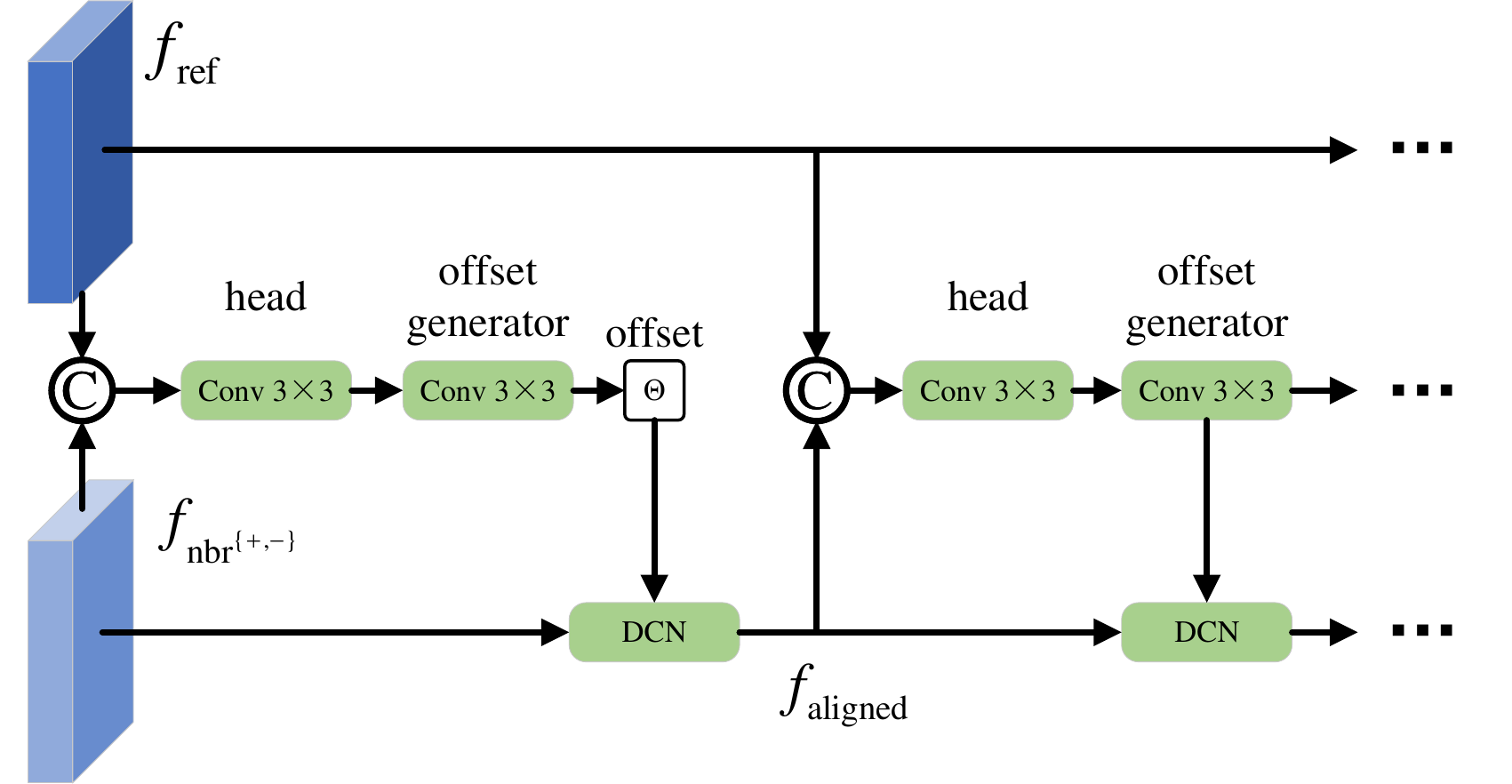}
	\end{center}
	\caption{In the architecture of cascading alignment, `$\mathrm{C}$' denotes concatenation. The alignment process will iterate 4 times (\emph{i.e.}, $L=4$), the initial aligned feature $f_{\mathrm{aligned}}^{0}$ is $f_{\mathrm{nbr}^{\left\{+, -\right\}}}$. The output of the previous alignment module will be concatenated with $f_{ref}$ and fed into the next alignment module.}
	\label{fig:casalign}
\end{figure}

\textbf{Cascading alignment}. Taking advantage of the deformable convolution network (DCN) \cite{DCN, DCNV2}, we use an adaptive deformable kernel to achieve feature-level alignment. Different from optical flow-based methods, the alignment is applied on feature of each frame, denoted by $f_{\mathrm{ref}}$ and $f_{\mathrm{nbr}^{\left\{+, -\right\}}}$. As such, image warping is not required. The alignment module consists of multiple alignment operations organized in a cascaded manner, as shown in Fig. \ref{fig:casalign}. We first take the feature of the center moment $f_{\mathrm{ref}}$ as a reference feature to align each neighboring feature $f_{\mathrm{nbr}^{sign}}$ $(sign \in \left\{+, -\right\})$, and the feature-level offset $\Theta$ between the reference feature and neighboring feature is estimated. In this process, the reference and neighboring feature are concatenated before the transformation with a typical convolution layer $\tau_{\theta}(\cdot)$ and a light-weight offset generator $\phi_{\theta}(\cdot)$:
\begin{equation}\label{eq:casalign:1}
	\Theta^{l}=\phi_{\theta}^{l}(\tau_{\theta}^{l}(\left[f_{\mathrm{ref}}, f_{\mathrm{aligned}}^{l-1}\right])), l \in \left\{1, ..., L\right\},
\end{equation}
where $l$ denotes cascaded layer in depth, $\theta$ is the learnable parameters, and $\left[\cdot, \cdot\right]$ represents the concatenation operation. In addition, we have $f_{\mathrm{aligned}}^{0}=f_{\mathrm{nbr}^{\left\{+, -\right\}}}$. For simplicity, we omit the superscript $l$ for $\Theta^{l}$, and then $\Theta$ can be formulated as:
\begin{equation}\label{eq:casalign:2}
	\Theta=\left\{\Delta p_{n}\mid n=1,...,\left \| \mathcal{R}\right \| \right\}.
\end{equation}
Here, $\Theta$ refers to learnable offsets for the convolution kernels of sampling locations. For instance, the regular grid of $3\times 3$ convolution kernel has 9 sampling locations, where $\mathcal{R}=\{(-1, -1), (-1, 0), ..., (0, 1), (1, 1)\}$. Then, DCN is applied to align the neighboring features to the reference feature with the guidance of the offset map $\Theta$:
\begin{equation}\label{eq:casalign:3}
	\begin{aligned}
		f_{\mathrm{aligned}}^{l}&=\mathrm{DCN}(f_{\mathrm{aligned}}^{l-1}, \Theta^{l}) \\
		&=\sum_{p_{n}\in\mathcal{R}}w(p_{n})f_{\mathrm{aligned}}^{l-1}(p_{0}+p_{n}+\Delta p_{n}),
	\end{aligned}
\end{equation}
where $f_{\mathrm{aligned}}^{l-1}$ is aligned feature at the $(l-1)$-th stage ($l\in \{1, ..., L\}$). $w$ and $p_{0}$ denote the weight and sampling locations of a convolutional kernel. As the $\Theta$ is fractional, the sampling operation is implemented by using bilinear interpolation as in \cite{DCN}.

Cascading alignment module consists of several regular and deformable convolutional layers. For the offset generation phase, it concatenates $f_{\mathrm{ref}}$ and $f_{\mathrm{aligned}}^{l}$ ($l \in \{0, ..., L\}$) and then uses a $3\times 3$ convolution layer to reduce the channel number of the concatenated feature map and follow the same kernel size convolution layer to predict the sampling parameters. Finally, the aligned feature $f_{\mathrm{aligned}}^{l}$ is obtained from $f_{\mathrm{aligned}}^{l-1}$ and $\Theta^{l}$ based on the DCN, where $l\in\{1, ..., L\}$. We use a four-stage cascaded structure, \emph{i.e.}, $L=4$, which gradually refines the coarsely aligned features. The cascading alignment module in such a coarse-to-fine manner improves the alignment at the feature level. 

\textbf{Temporal-spatial compensation}. Inspired by the CBAM \cite{CBAM}, we propose temporal-spatial compensation (TSC) to exploit both temporal and spatial-wise attention on each frame. We focus on `where' is an informative part supported by reference feature $f_{\mathrm{ref}}$. In addition, we employ a pyramid structure to increase the attention receptive field due to unequally information of different aligned induced by occlusion and blur. 

\begin{figure}[t]
	\begin{center}
		\includegraphics[width=0.9\linewidth]{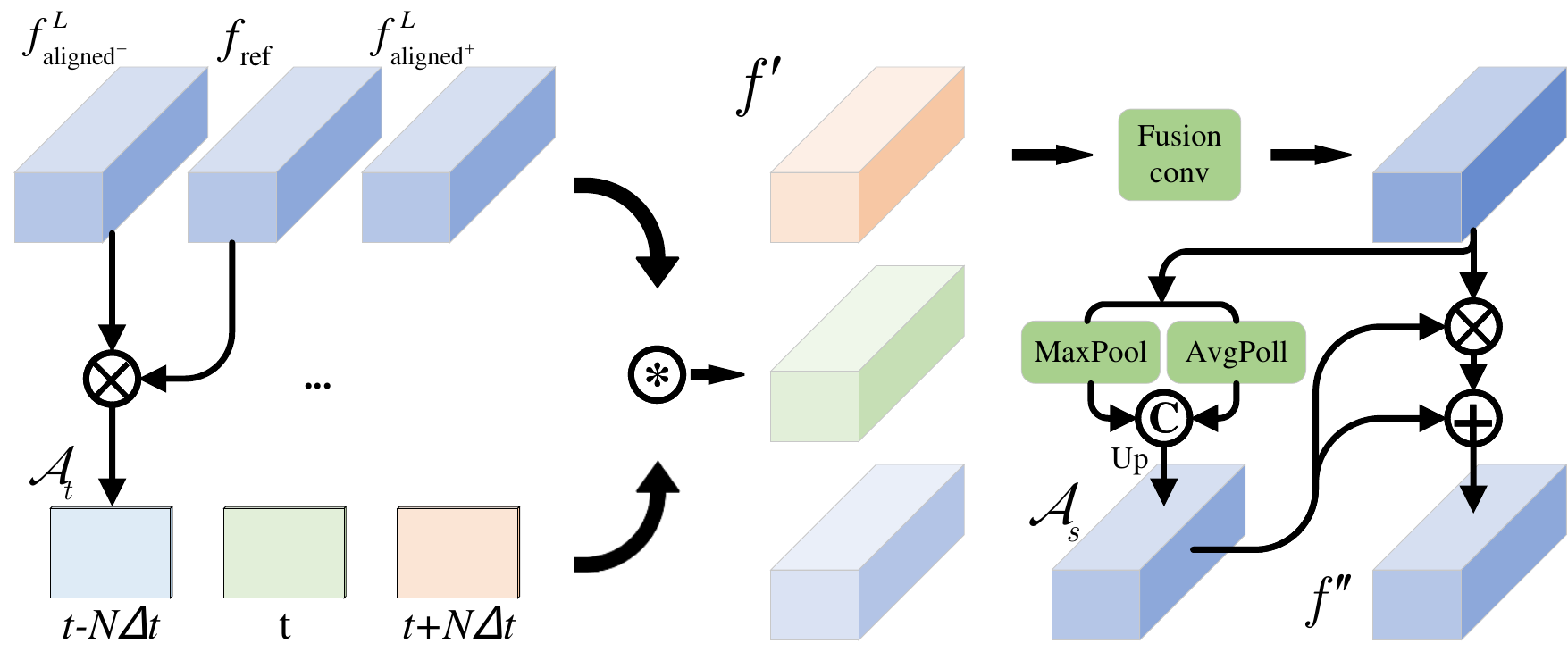}
	\end{center}
	\caption{The computational graph of temporal-spatial compensation operation. `$\mathrm{C}$' represents the concatenation. `$\mathrm{MaxPool}$' and `$\mathrm{AvgPool}$' are max and average pooling layers with stride 2.}
	\label{fig:tsc}
\end{figure}

Given the reference feature $f_{\mathrm{ref}}\in \mathbb{R}^{\frac{w}{4}\times\frac{h}{4}\times C}$ and concatenated feature $[f_{\mathrm{aligned}^{-}}^{L}, f_{\mathrm{ref}}, f_{\mathrm{aligned}^{+}}^{L}] \in \mathbb{R}^{\frac{w}{4}\times\frac{h}{4}\times (2N+1)C}$ as inputs, the TSC sequentially infers $2N+1$ 2D temporal attention map $\mathcal{A}_{t}\in\mathbb{R}^{\frac{w}{4}\times\frac{h}{4}\times 1}$  and a spatial attention $\mathcal{A}_{s}\in\mathbb{R}^{\frac{w}{4}\times\frac{h}{4}\times C}$, where $[\cdot, \cdot]$ denotes concatenate operation. The overall fusion process can be summarized as:
\begin{equation}\label{eq:tsc:1}
	\begin{matrix}
		\begin{aligned}
			{f}'&=\mathcal{A}_{t}([f_{\mathrm{aligned}^{-}}^{L}, f_{\mathrm{ref}}, f_{\mathrm{aligned}^{+}}^{L}]\mid f_{\mathrm{ref}})\ast\\
			&[f_{\mathrm{aligned}^{-}}^{L}, f_{\mathrm{ref}}, f_{\mathrm{aligned}^{+}}^{L}]\\ 
			{f}''&=\mathcal{A}_{s}({f}')\ast{f}'+\delta_{\theta}(\mathcal{A}_{s}({f}')),
		\end{aligned}
	\end{matrix}
\end{equation}
where `$\ast$' denotes the element-wise multiplication, and $\delta_{\theta}$ is a light-weight encoder with two $3\times 3$ convolutional layers followed by a ReLU activation. The temporal attention $\mathcal{A}_{t}(\cdot)$ is implemented by calculating the similarity distance:
\begin{equation}\label{eq:tsc:2}
	\mathcal{A}_{t}=\sigma(\mathrm{Sum}(\phi(f)^{\top}\cdot\psi(f_{ref})))
\end{equation}
where $f\in\left\{f_{\mathrm{aligned}^{-}}^{L}, f_{\mathrm{ref}}, f_{\mathrm{aligned}^{+}}^{L}\right\}$. $\phi(\cdot)$ and $\psi(\cdot)$ are achieved with simple convolution layers. $\mathrm{Sum}(\cdot)$ and $\sigma(\cdot)$ denote channel-wise summation and sigmoid activation function. The spatial attention $\mathcal{A}_{s}$ employs a pyramid design to increase the attention receptive field, $\mathcal{A}_{s}$ can be obtained by:
\begin{equation}\label{eq:tsc:3}
	\begin{matrix}
		\begin{aligned}
	\mathcal{A}_{s}=\mathrm{Up}(\theta_{2}([\mathrm{MaxPool}(\theta_{1}({f}')),\\ \mathrm{AvgPool}(\theta_{1}({f}'))]))+{f}'
		\end{aligned}
	\end{matrix}
\end{equation}
where $\theta_{1}(\cdot)$ and $\theta_{2}(\cdot)$ are the typical convolution layer with dimensional reductions as original input channels. $\mathrm{MaxPool}(\cdot)$ and $\mathrm{AvgPool}(\cdot)$ represent max and average pooling layers with stride 2. Note that we add embedding spatial attention $\mathcal{A}_{s}$ to spatial enhanced features to enhance losing important information on the regions with attention values close to zero. ${f}''\in\mathbb{R}^{\frac{w}{4}\times\frac{h}{4}\times C}$ is the final refined output. Fig. \ref{fig:tsc} depicts the computation process of each attention map.

\textbf{Segmentation}. The segmentation module consists of a residual block (with 64 filters) and a $1\times 1$ convolutional layer (with one filter and sigmoid) for the final segmentation prediction based on the last output ${f}''$ of TSC.

\subsection{Joint training strategy}\label{subsec:35}

\begin{figure}[htbp]
	\begin{center}
		\includegraphics[width=0.95\linewidth]{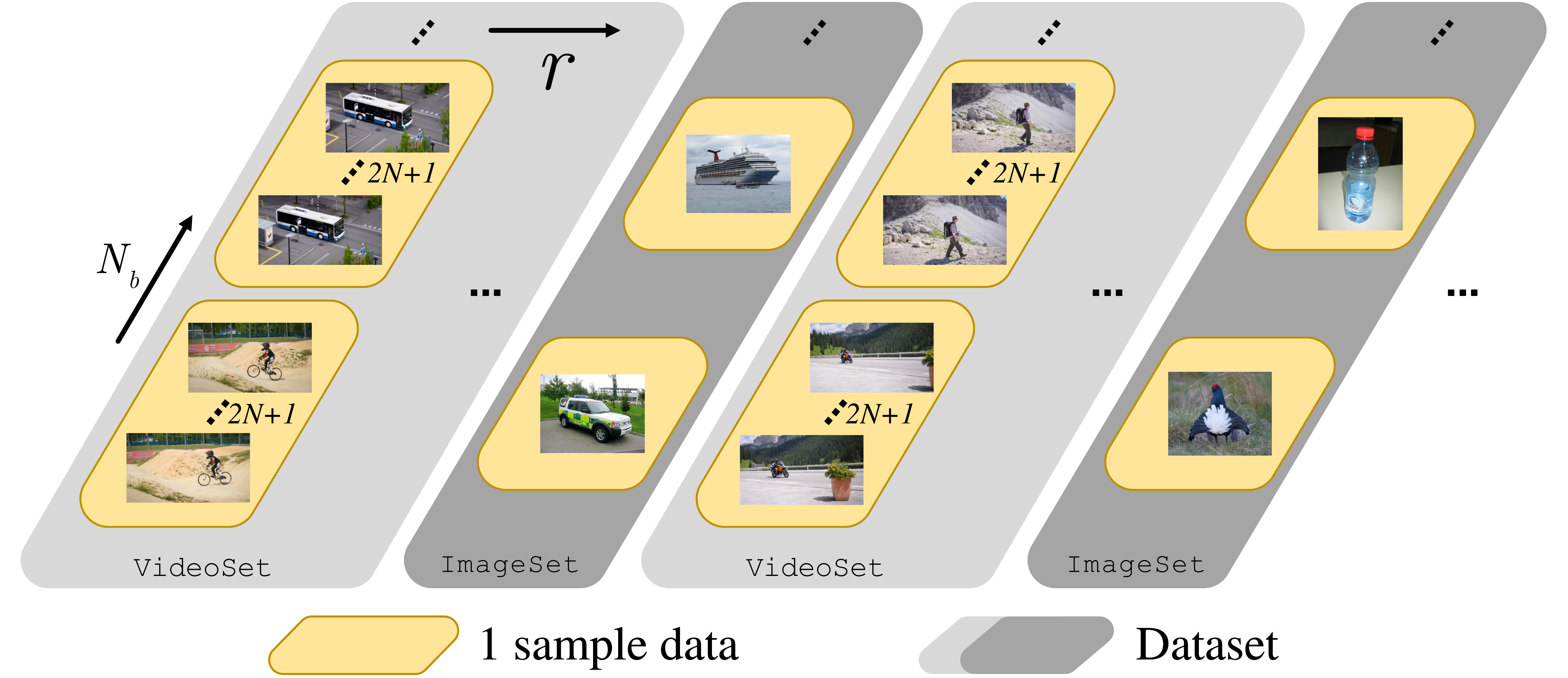}
	\end{center}
	\caption{Demo of the joint training strategy. Sampling $N_{b}$ video sequence samples from the \texttt{VideoSet} create one batch video sample, and likewise, sampling $N_{b}$ image samples from the \texttt{ImageSet} constitutes one batch image sample. One batch image sample is matching with $r$ batches video samples constitute an iteration for joint training. Here, $r$ is computed by Eq. \ref{eq:JointTr:1}.}
	\label{fig:JointTr}
\end{figure}

A possible issue of the existing methods is that the temporal consistency is not maintained across the whole of the video sequence in the presence of visual similar objects or surroundings. The reason is that these methods focus on mining the correlation (global dependence) between consecutive frames, but pay less attention to the saliency object (local dependence), leading to inaccurate segmentation results. To alleviate this problem, we use the SOD and UVOS datasets to jointly train our model for the UVOS task. The efficient joint training strategy is summarized in Fig. \ref{fig:JointTr}.

The SOD and UVOS datasets are divided into two groups: \texttt{ImageSet} and \texttt{VideoSet}. The \texttt{ImageSet} obtained by sampling from $\textit{DUTS}$ \cite{DUTS} dataset consists of each individual image, and the \texttt{VideoSet} is temporal related consecutive frames getting from the $\textit{DAVIS}_{\textit{16}}$ \cite{DAVIS2016} dataset. $N_{b}$ images or video clips where one clip contains $2N+1$ frames are randomly sampled in each mini-batch from $\textit{DUTS}$ or $\textit{DAVIS}_{\textit{16}}$ datasets, respectively. One mini-batch \texttt{ImageSet} is repeated after $r$ mini-batches \texttt{VideoSet}. The ratio $r$ is given by
\begin{equation}\label{eq:JointTr:1}
	r=\left \lfloor \frac{\mathrm{len}(\texttt{VideoSet})}{(\left \lfloor \mathrm{len}(\texttt{ImageSet}) / N_{b} \right \rfloor \times N_{b})} \right \rfloor,
\end{equation}
where $\left \lfloor \cdot \right \rfloor$ denotes the floor function, and $\mathrm{len}(\texttt{dataset})$ is return the size of the $\texttt{dataset}$. The role of the \texttt{ImageSet} is that the saliency object constraint is enforced to make the problem tractable. Through the above steps, the IMCNet can effectively improve the final segmentation results, as presented in section \ref{sec:exp}.

\subsection{Loss functions}\label{subsec:36}

The loss functions in \cite{BASNet} are adopted to jointly measure the prediction in the pixel level by binary cross-entropy loss \cite{BCELOSS}, in the patch level by SSIM loss \cite{SSIMLOSS}, as well as in the region level by IoU loss \cite{IoULOSS}:
\begin{equation}\label{eq:lf:1}
	\mathscr{L}(\hat{M}, M)=\mathscr{L}_{bce}(\hat{M}, M)+\mathscr{L}_{ssim}(\hat{M}, M)+\mathscr{L}_{iou}(\hat{M}, M),
\end{equation}
where $\hat{M}$ denotes the segmentation prediction and $M$ refers to the binary ground-truth.

Given consecutive frames $\{I_{i} \in \mathbb{R}^{w \times h \times 3}\}_{i=t-N\Delta}^{t+N\Delta}$ from a video sequence, our IMCNet predicts a final segmentation mask $\hat{M}_{t}\in\left\{0, 1\right\}^{w\times h}$ and intermediate segmentation side-outputs $\left\{\hat{M}_{i}^{l}\in\left\{0, 1\right\}^{w\times h}\right\}_{i=t-N\Delta t}^{t+N\Delta t}$ through $l$-th ($l\in\left\{2, ..., L\right\}$) level of APM, and our APM is deeply supervised with the last four stages, \emph{i.e.} $L=5$. The total loss is formulated as:
\begin{equation}\label{eq:lf:2}
	\begin{aligned}
		\mathscr{L}_{total}&=\mathscr{L}_{final}(\hat{M}_{t}, M_{t})+\mathscr{L}_{side}(\hat{M}_{i}^{l}, M_{i})\\
		&=\mathscr{L}(\hat{M}_{t}, M_{t})+\sum_{i=t-N\Delta t}^{t+N\Delta t}\sum_{l=2}^{L}\mathscr{L}(\hat{M}_{i}^{l}, M_{i}).
	\end{aligned}
\end{equation}

\section{Experiments}\label{sec:exp}

Extensive experimental results are presented to evaluate the proposed IMCNet. 

\subsection{Datasets and evaluation metrics}

To test the performance of our method, we carry out comprehensive experiments on two UVOS datasets: 

$\textit{DAVIS}_{\textit{16}}$ \cite{DAVIS2016} is currently the most popular VOS benchmark, which consists of 50 high-quality video sequences (30 videos for training and 20 for testing). Each frame is densely annotated pixel-wise ground-truth for the foreground objects. We train our IMCNet on the training set and evaluate on the validation set. For quantitative evaluation, we adopt two standard metrics suggested by \cite{DAVIS2016}, namely region similarity $\mathcal{J}$, which is the intersection-over-union of the prediction and ground-truth, boundary accuracy $\mathcal{F}$, which measures the accuracy of the predicted mask boundaries. These measures can be averaged to give an overall $\mathcal{J}\&\mathcal{F}$ score.

$\textit{YouTube-Objects}$ \cite{YouTubeObj} contains 126 web video sequences that assign 10 semantic object categories with more than 20,000 frames in total. All the video sequences are used for performance evaluation. Following $\textit{YouTube-Objects}$'s evaluation protocol, we use the region similarity $\mathcal{J}$ to measure the segmentation performance.

\subsection{Implementation details}\label{subsec:imple}

\subsubsection{Detailed network architecture}

The backbone of IMCNet is the ResNet101 \cite{ResNet}, for each input frame of size $480\times 480 \times 3$, the frame is a down-sample to the size of $\left\{120, 60, 30, 15\right\}$ in the last four layers of ResNet101. The network takes three consecutive frames (\emph{i.e.}, $N=1$) with an interval of four frames (\emph{i.e.}, $\Delta t=4$) as inputs unless otherwise specified. The output size of our IMCNet is $480\times480\times1$, for the UVOS task, we use the bilinear interpolate algorithm to restore the original size of the input. For the ACM, we implement weight matrices $P^{\top}$ and $Q^{\top}$ in Eq. \ref{eq:ac:2} using two $1\times1$ convolutional layers with 64 kernels, respectively. The channel size $C$ in the TSC module is set to 64.

\subsubsection{Training settings}

The training data consists of three parts: 1) all training data from the $\textit{DAVIS}_{\textit{16}}$ \cite{DAVIS2016}, including 30 videos with about 2K frames; 2) a subset of the training set of $\textit{YouTube-VOS}$ \cite{YOUTUBEVOS} selected 18K frames, which is obtained by sampling images containing a single object per sequence; 3) $\textit{DUTS-TR}$ which is the training set of $\textit{DUTS}$ \cite{DUTS} has more than 10K images. The training process is divided into two stages. Stage 1: we first pre-train our network for 200K iterations on a subset of $\textit{YouTube-VOS}$. During this training period, the entire network is trained using Adam \cite{Adam} optimizer ($\beta_{1}=0.9$ and $\beta_{2}=0.999$) with a learning rate of $10^{-6}$ for the FEM, $10^{-5}$ for the ACM and APM decoder, and the MCM is set as $10^{-4}$. The batch size is set to $8$, and the weight decay is $0$. Stage 2: we fine-tune the entire network on the training set of $\textit{DAVIS}_{\textit{16}}$ and $\textit{DUTS}$ with our joint training strategy. In this stage, the Adam optimizer is used with an initial learning rate of $10^{-7}$, $10^{-6}$, and $10^{-5}$ for each of the above-mention modules, respectively. We train our network for 155K iterations in the second training stage. Data augmentation (\emph{e.g.}, scaling, flipping, and rotation) is also adopted for both image and video data. Our IMCNet is implemented in PyTorch \cite{PyTorch}. All experiments and analyses are conducted on an NVIDIA TITAN RTX GPU, and the overall training time is about 72 hours.

\subsubsection{Test settings}

On the test time, we resize the input frame to $480\times480\times3$, and feed three consecutive frames with an interval of four frames into the network for segmentation. The segmentation mask $\hat{M}_{t}$ of the center frame $I_{t}$ is obtained from the IMCNet. We follow the common protocol used in existing works \cite{ADNet, DFNet} and employ multi-scale and mirrored inputs to enhance the final segmentation mask without CRFs and instance pruning.

\subsubsection{Runtime}

During testing, the forward estimation of our IMCNet takes around 0.05s per frame, while the post-processing takes about 0.2s.

\subsection{Ablation study}\label{subsec:abla}

To demonstrate the influence of each component in the IMCNet, we perform an ablation study on the test set of $\textit{DAVIS}_{\textit{16}}$. The evaluation criterion is mean region similarity ($\mathcal{J}$) and mean boundary accuracy ($\mathcal{F}$).

\subsubsection{Effectiveness of ACM}

To demonstrate the effectiveness of the ACM, we gradually remove the affinity computing and the key encoder process in our model, denoted as \emph{w/o.} ACM and \emph{w/o.} key encoder, respectively. It means that the features in the last convolution stage of the FEM are directly fed into the APM to achieve segmentation results. The results can be referred in the sub-table named \textit{ACM} in Table \ref{tab:AblaStudy}. From the results, we can observe that the performance of the variant without ACM (affinity computing + key encoder) is $-2.7\%$ lower than our full model (with affinity computing + key encoder) in terms of mean $\mathcal{J}$, and $-3.1\%$ lower on mean $\mathcal{F}$, respectively. In other words, ACM is key to improving the performance of our proposed method ($2.7\%$ and $3.1\%$ on mean $\mathcal{J}$ and mean $\mathcal{F}$, respectively). On this basis, we only introduced the affinity computing process, the performance of the variant without the key encoder is improved by $0.6\%$ and $0.7\%$ in terms of mean $\mathcal{J}$ and mean $\mathcal{F}$ higher than the variant without ACM. This indicates that affinity computing improves the performance of the model ($0.6\%$ and $0.7\%$ on mean $\mathcal{J}$ and mean $\mathcal{F}$, respectively). After that, we can observe that the performance of our full model (with affinity computing and key encoder) is $2.1\%$ and $2.4\%$ higher than the variant without key encoder on mean $\mathcal{J}$ and $\mathcal{F}$, respectively. This validates the effectiveness of our key encoder that eliminates the redundancy by only focusing on the features related to the primary object(s). Meanwhile, our model can address the UVOS task by capturing global dependence from the multi-frames, which verifies the effectiveness of ACM.

\begin{table}[htb]
	\scriptsize
%	\tiny
	\caption{Ablation study of IMCNet on the $\textit{DAVIS}_{\textit{16}}$ dataset with different variants and training strategies, measured by the mean $\mathcal{J}$ and mean $\mathcal{F}$. See \S \ref{subsec:abla} for details.}
	\begin{center}
		\resizebox{1.0\columnwidth}{!}{
			\begin{tabular}{c|cc|cc} \toprule
				\multirow{2}[0]{*}{Network Variant} & \multicolumn{4}{c}{$\textit{DAVIS}_{\textit{16}}$ \texttt{val}} \\
				& Mean $\mathcal{J}\uparrow$ & $\Delta\mathcal{J}$ & Mean $\mathcal{F}\uparrow$ & $\Delta\mathcal{F}$ \\ \midrule \midrule
				\multicolumn{5}{c}{\textit{ACM}} \\ \midrule
				\emph{w/o.} ACM &   80.0    &   -2.7    &    78.0   & -3.1 \\
				\emph{w/o.} key encoder (only affinity computing) &   80.6    &   -2.1    &   78.7    & -2.4 \\ \midrule \midrule
				\multicolumn{5}{c}{\textit{APM}} \\ \midrule
				\emph{w/o.} APM   &   80.4    &   -2.3    &   80.1    & -1.0 \\ \midrule \midrule
				\multicolumn{5}{c}{\textit{Cascading Alignment}} \\ \midrule
				regular &   81.5    &    -1.2   &   80.0    & -1.1 \\ 
				$l=1$ &   79.6    &    -3.1   &   78.4    & -2.7 \\ 
				$l=2$ &   80.7    &    -2.0   &   79.3    & -1.8 \\ 
				$l=3$ &   81.4    &    -1.3   &   80.1    & -1.0 \\ 
				$l=4$ &   \textbf{82.7}    &    -   &   \textbf{81.1}    & - \\ 
				$l=5$ &   82.0    &    -0.7   &   81.0    & -0.1 \\ \midrule \midrule
				\multicolumn{5}{c}{\textit{TSC}} \\ \midrule
				\emph{w/o.} TSC (Conv) &    80.6   &   -2.1    &   79.3    & -1.8 \\ \midrule \midrule
				\multicolumn{5}{c}{\textit{Training Strategy}} \\ \midrule
				Stage 1 &   76.3    &   -6.4    &   75.0    & -6.1 \\
				Stage 1 \& Stage 2 (\emph{w/o.} joint training) &   79.3    &    -3.4   &   78.4    & -2.7  \\
				Stage 1 \& Stage 2 (\emph{w/.} joint training)   &   \textbf{82.7}    &     -      &   \textbf{81.1}    & - \\ \bottomrule
			\end{tabular}%
		}
	\end{center}
	\label{tab:AblaStudy}%
\end{table}%

\subsubsection{Effectiveness of APM}

\begin{figure*}[htb]
	\begin{center}
		\includegraphics[width=0.9\linewidth]{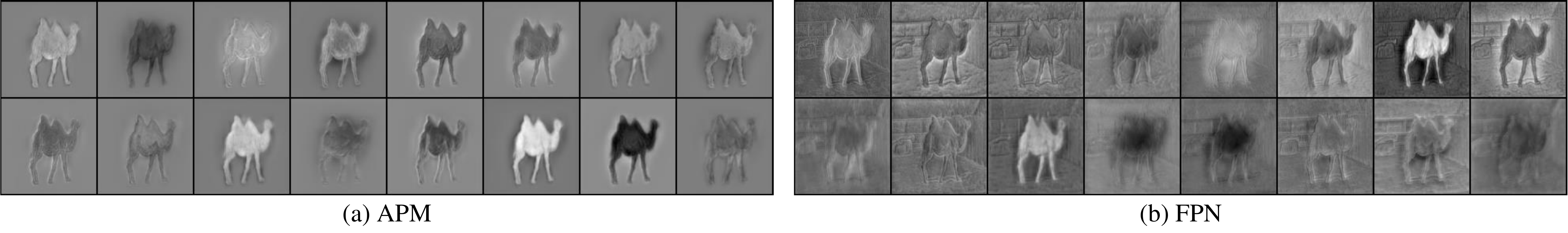}
	\end{center}
	\caption{Visualization of output feature maps with the decoder on \textit{camel} video sequences. The left subfigure (a) shows 16 channels features with the APM, and the right subfigure (b) shows features with the FPN. The feature map with the APM pays more attention to the foreground.}
	\label{fig:APMvsFPN}
\end{figure*}

The purpose of APM (Eq. \ref{eq:apm:1}) is to retain high-level features related to the primary object(s) during the top-down refinement. To verify such a design, we implement another network by replacing the APM with the FPN \cite{FPN} architecture in which the feature dimension is set as 64 following the same dimension as the APM, denoted as \emph{w/o.} APM. The results in Table \ref{tab:AblaStudy} (see \emph{w/o.} APM) demonstrate the superiority of APM. As shown in Fig. \ref{fig:APMvsFPN}, the variant, with the FPN architecture, is vulnerable to being distracted by irrelevant surroundings when top-down refining object details. Therefore, this variant leads to an obvious drop in performance.

\begin{figure*}[htbp]
	\begin{center}
		\includegraphics[width=0.95\linewidth]{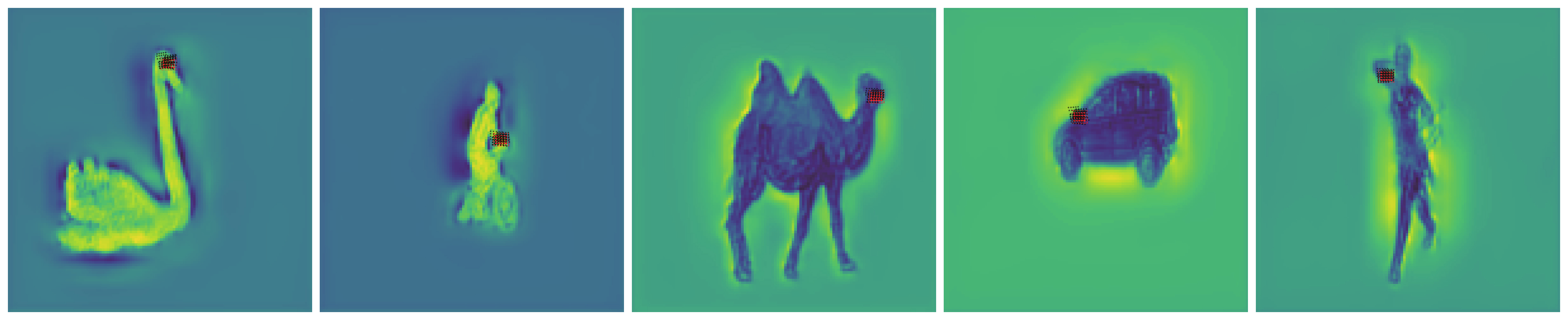}
	\end{center}
	\caption{Illustration of the sampling locations in cascading alignment. The red points in each image represent five adjacent $3\times3$ regular convolution kernels, and the receptive fields are plotted by a red dotted rectangle. Black points in each image indicate the learned sampling position for deformable convolution kernel, relative to the original position (red points).}
	\label{fig:DCN_Kernel}
\end{figure*}

\begin{figure*}[htb]
	\begin{center}
		\includegraphics[width=0.9\linewidth]{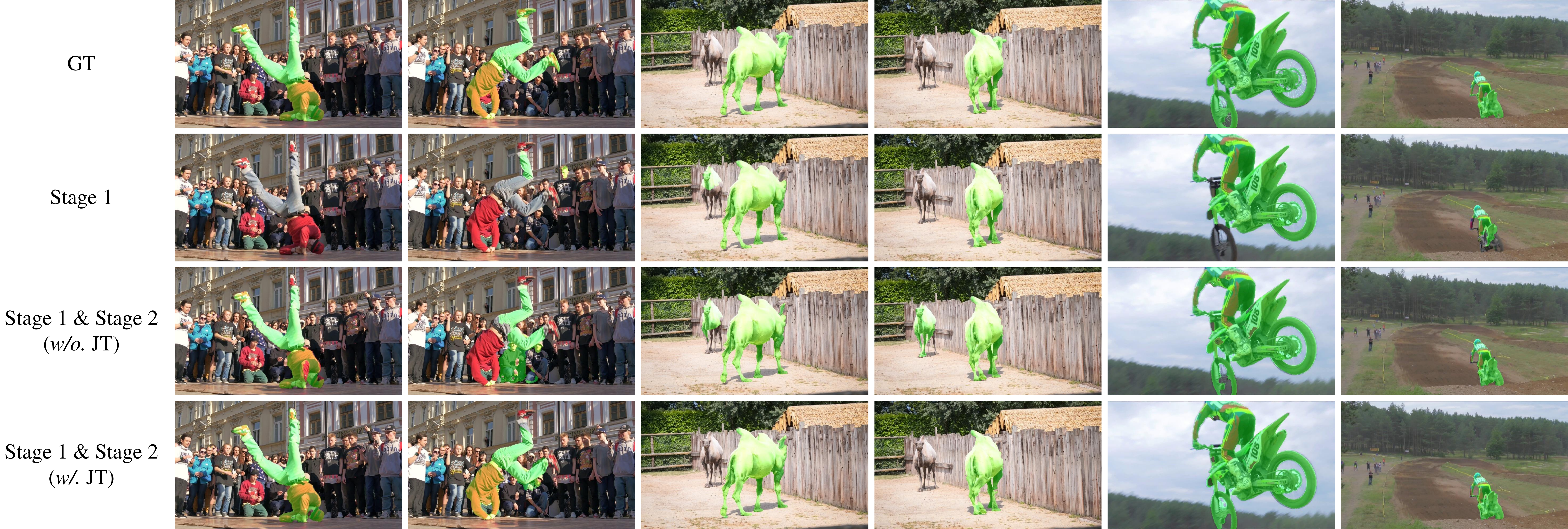}
	\end{center}
	\caption{Qualitative results on three sequences for different training strategies. From left to right: \textit{breakdance}, \textit{camel}, and \textit{motocross-jump} from the $\textit{DAVIS}_{\textit{16}}$. GT and JT denote ground-truth and joint training, respectively.}
	\label{fig:JT}
\end{figure*}

\subsubsection{Effectiveness of cascading alignment}

We study the effectiveness of the cascading alignment operation in Eq. \ref{eq:casalign:3} by comparing our whole model to one variant in which the deformable convolutional layer is replaced by a $3\times3$ regular convolutional kernel. We can draw the following conclusion from results in the sub-table named \textit{Cascading Alignment} in Table \ref{tab:AblaStudy}. The cascading alignment process can achieve an absolute gain of $1.2\%$ and $1.1\%$ in mean $\mathcal{F}$ and mean $\mathcal{J}$, respectively. Fig. \ref{fig:DCN_Kernel} shows the visualization of sampling positions on immediate features in cascading alignment. We use red points ($5^{2}=25$ points in each feature map) to represent the sampling positions of the regular convolution network, and black points represent the sampling locations of deformable convolution in the cascading alignment. We observe that the sampling positions in the regular convolution are fixed all over the feature map, while they in APM tend to adaptively adjust according to objects' shape and scale. The quantitative evidence of such adaptive deformation is provided in Table \ref{tab:AblaStudy}.

Moreover, we also evaluate the performance of the IMCNet with a different number of cascading layers of alignment. We can see from the results in \textit{Cascading Alignment} of Table \ref{tab:AblaStudy} that the performance gradually improves as the number of cascading layer $l$ increases, reaching the best at $l=4$. Therefore, the default value of $l$ is set as 4 for the cascading alignment process.

\subsubsection{Effectiveness of TSC}

To verify the effect of the TSC, we only add a bridge stage, which is implemented by a $3\times3$ convolutional layer, between the cascading alignment and segmentation module. As shown in Table \ref{tab:AblaStudy}, in the sub-table \textit{TSC}, the variant without TSC suffers from a significant performance drop ($-2.2\%$ in mean $\mathcal{J}$ and $-1.8\%$ in mean $\mathcal{F}$), which verifies the effectiveness of TSC.

\subsubsection{Comparison of different training strategies}

In order to enhance the local dependence during the convergence of training, we introduce a joint training strategy to capture intra-frame discriminability. The joint training strategy helps guarantee our model focuses on mining both local and global dependence. Our IMCNet is trained with our proposed joint training strategy in Stage 2. To investigate the efficacy of joint training strategy on the UVOS task, we train our model with only Stage 1 and Stage 1\&Stage 2 \emph{w/o.} joint training. Its comparison results in \textit{Training Strategy} of Table \ref{tab:AblaStudy} demonstrate the effectiveness of such a joint training method. In Stage 1, our model is trained with a subset of \textit{YouTube-VOS} (annotated one objects mask), it is not reliable for characterizing objects surrounded by cluttered background (see \textit{breakdance} in the second row of Fig. \ref{fig:JT}), leading to poor generalization. Especially, it easily fails in the presence of fast motion (in \textit{breakdance} and \textit{motocross-jump} sequences). However, for intra-frame discriminability, it is better than Stage 1 \& Stage 2 without joint training, as shown \textit{camel} in the second and third rows of Fig. \ref{fig:JT}. The IMCNet which is trained with Stage 1 \& Stage 2 (\emph{w/o.} joint training) is sensitive to similar surroundings and distracting backgrounds, while joint training strategy effectively improves the intra-frame discriminability (see \textit{breakdance} and \textit{camel} in the fourth row of Fig. \ref{fig:JT}).

\subsection{Influence of key parameters}\label{subsec:key}

In this section, we analyze the influence of key parameters in our IMCNet, including the frame number $N$, and interval step $\Delta t$, for input frames on a densely annotated $\textit{DAVIS}_{\textit{16}}$ dataset. The networks are evaluated using mean region similarity ($\mathcal{J}$) and mean boundary accuracy ($\mathcal{F}$).

\subsubsection{Frame number}

Our IMCNet simultaneously takes $2N+1$ frames as inputs and generates the segmentation mask of the center frame ($i=t$). It is of interest to assess the influence of the number of input frames $2N+1$ ($N\in[1, 3]$) on the final performance. Table \ref{tab:KeyPara} shows the results for this. From the results, we can see that the performance can deteriorate as $N$ increases. When $N$ is larger, it means that the information of frames which are far from the center frame is also considered, which may lead to noisy information. Based on this analysis, the IMCNet achieves the best performance of $82.7\%$ in mean $\mathcal{J}$ and $81.1\%$ in mean $\mathcal{F}$ when $N=1$ on the $\textit{DAVIS}_{\textit{16}}$ dataset.

\begin{table}[t]
	\tiny
	\caption{Comparisons with key parameters (frame number and step) of IMCNet on the $\textit{DAVIS}_{\textit{16}}$ dataset, measured by the mean $\mathcal{J}$ and mean $\mathcal{F}$. See \S \ref{subsec:key} for details.}
	\begin{center}
		\resizebox{1.0\columnwidth}{!}{
			\begin{tabular}{c|cc|cc} \toprule
				\multirow{2}[0]{*}{Network Variant} & \multicolumn{4}{c}{$\textit{DAVIS}_{\textit{16}}$ \texttt{val}} \\
				& Mean $\mathcal{J}\uparrow$ & $\Delta\mathcal{J}$ & Mean $\mathcal{F}\uparrow$ & $\Delta\mathcal{J}$ \\ \midrule \midrule
				\multicolumn{5}{c}{\textit{Frame Number} ($2N+1$)} \\ \midrule
				$5$   &   80.2    &   -2.5    &   78.5    & -2.6 \\
				$7$  &    79.7   &   -3.0    &    78.0   & -3.1 \\ \midrule \midrule
				\multicolumn{5}{c}{\textit{Step} ($\Delta t$)} \\ \midrule
				$1$     &   82.6    &    -0.1   &   80.8    &  -0.3 \\
				$2$     &    82.0   &    -0.7   &   80.2    &  -0.9 \\
				$4$     &   \textbf{82.7}    &     -  &   \textbf{81.1}    &  - \\
				$6$     &   81.9    &     -0.8  &   80.5    &  -0.6 \\ \bottomrule
			\end{tabular}%
		}
	\end{center}
	\label{tab:KeyPara}%
\end{table}%

\begin{table}[htbp]
	\centering
	\tiny
	\caption{Attribute-based ablation study on the $\textit{DAVIS}_{\textit{16}}$ dataset. We compare the mean $\mathcal{J}$ of different frame interval step $\Delta t$ under various attributes. The maximum and minimum results are marked in \textcolor[rgb]{1, 0, 0}{Red} and \textcolor[rgb]{0, 0, 1}{Blue}. $\Delta\mathcal{J}$ is the difference between the maximum and minimum values.}
	\resizebox{1.0\columnwidth}{!}{
		\begin{tabular}{c|cccc|c}\toprule
			\multirow{3}[2]{*}{Category} & \multicolumn{4}{c|}{Mean $\mathcal{J}$}         & \multirow{3}[2]{*}{$\Delta\mathcal{J}$} \\
			\cmidrule{2-5}          & \multicolumn{4}{c|}{\textit{Step}}      &  \\
			& 1     & 2     & 4     & 6     &  \\ \midrule \midrule
			AC    & \textcolor[rgb]{ 1,  0,  0}{84.0} & \textcolor[rgb]{  0,  0,  1}{82.2} & 83.9  & 83.0  &1.8\\
			BC    & 81.6  & 81.8  & \textcolor[rgb]{ 1,  0,  0}{81.9} & \textcolor[rgb]{  0,  0,  1}{81.5} & 0.4 \\
			CS    & 84.3  & 84.4  & \textcolor[rgb]{ 1,  0,  0}{84.5} & \textcolor[rgb]{  0,  0,  1}{84.0} & 0.5\\
			DB    & \textcolor[rgb]{ 1,  0,  0}{69.7} & \textcolor[rgb]{  0,  0,  1}{63.7} & 68.5  & 66.3  & 6.0 \\
			DEF   & 80.6  & \textcolor[rgb]{  0,  0,  1}{79.8} & \textcolor[rgb]{ 1,  0,  0}{81.0} & 80.5  & 1.2 \\
			EA    & 78.5  & \textcolor[rgb]{ 1,  0,  0}{78.8} & 78.6  & \textcolor[rgb]{  0,  0,  1}{77.4} & 1.4 \\
			FM    & 78.6  & \textcolor[rgb]{  0,  0,  1}{76.7} & \textcolor[rgb]{ 1,  0,  0}{78.6} & 77.4  & 1.9 \\
			HO    & 78.1  & \textcolor[rgb]{  0,  0,  1}{77.3} & \textcolor[rgb]{ 1,  0,  0}{78.3} & 77.4  & 1.0\\
			IO    & \textcolor[rgb]{  0,  0,  1}{78.3} & 78.7  & \textcolor[rgb]{ 1,  0,  0}{78.7} & 78.3  & 0.4 \\
			LR    & \textcolor[rgb]{ 1,  0,  0}{80.1} & 80.1  & 80.0  & \textcolor[rgb]{  0,  0,  1}{78.8} & 1.3 \\
			MB    & 77.8  & \textcolor[rgb]{  0,  0,  1}{76.6} & \textcolor[rgb]{ 1,  0,  0}{78.3} & 77.6  & 1.7 \\
			OCC   & 76.7  & 76.9  & \textcolor[rgb]{ 1,  0,  0}{77.2} & \textcolor[rgb]{  0,  0,  1}{76.7} & 0.5 \\
			OV    & 81.2  & \textcolor[rgb]{  0,  0,  1}{78.0} & \textcolor[rgb]{ 1,  0,  0}{81.3} & 80.7  & 3.3 \\
			SC    & \textcolor[rgb]{  0,  0,  1}{71.9} & 72.5  & \textcolor[rgb]{ 1,  0,  0}{72.6} & 72.1  & 0.7\\
			SV    & 78.1  & \textcolor[rgb]{ 1,  0,  0}{78.6} & 78.4  & \textcolor[rgb]{  0,  0,  1}{77.3} & 1.3\\ \bottomrule
		\end{tabular}%
	}
	\label{tab:StepAttr}%
\end{table}%

\subsubsection{Step $\Delta t$}

Another key parameter is the step of frames $\Delta t$, which decides sequential frames as input in an ordered manner is selected at a fixed-length frame interval $\Delta t$. $\Delta t=1$ represents selecting three consecutive adjacent video frames in input frames. \textit{Step} in Table \ref{tab:KeyPara} reports the mean $\mathcal{J}$ and mean $\mathcal{F}$ as a function of the frame interval $\Delta t$. We can see that when $\Delta t$ increase, the mean $\mathcal{J}$ and mean $\mathcal{F}$ first increase and then decrease.

Moreover, Table \ref{tab:StepAttr} illustrates the performance comparison of different $\Delta t$ under various video attributes of $\textit{DAVIS}_{\textit{16}}$, including low resolution (LR), scale variation (SV), shape complexity (SC), fast motion (FM), camera-shake (CS), interacting objects (IO), dynamic background (DB), motion blur (MB), deformation (DEF), occlusion (OCC), heterogeneous object (HO), edge ambiguity (EA), out-of-view (OV), appearance change (AC), and background cluttering (BC). IMCNet with $\Delta t=4$ has the best performance under most attributes. As a result, in the presence of appearance change (AC), dynamic background (DB), and low resolution (LR), the model with $\Delta t=1$ is the most robust due to the primary object(s) undergoing huge appearance change, scale variation, and it is difficult to capture the similarity between multi-frames when $\Delta t$ is larger. We can reach the same conclusion from $\Delta\mathcal{J}$ in Table \ref{tab:StepAttr} that dynamic background (DB) and out-of-view (OV) have the greatest influence on interval step $\Delta t$.

\subsection{Comparison with state-of-the-arts}

We compare our proposed IMCNet with the state-of-the-art methods in two densely annotated video segmentation datasets, \emph{i.e.}, $\textit{DAVIS}_{\textit{16}}$ \cite{DAVIS2016} and $\textit{YouTube-Objects}$ \cite{YouTubeObj}. We apply the training strategy mentioned in \S \ref{subsec:imple}. Input frame numbers and step $\Delta t$ are set as 3 and 4.

\begin{table*}[htb]
	\tiny
	\centering
	\caption{Quantitative results on the test set of $\textit{DAVIS}_{\textit{16}}$, using the region similarity $\mathcal{J}$, boundary accuracy $\mathcal{F}$. The top three results are marked in \textcolor[rgb]{1, 0, 0}{Red}, \textcolor[rgb]{0, 1, 0}{Green}, and \textcolor[rgb]{0, 0, 1}{Blue}. We also report the input modality for each UVOS method in the second column. \textrm{OF}, \textrm{RGB} and \textrm{MF} represent optical flow, RGB image and multi-frames input in test time, respectively.}
	\begin{threeparttable}
		\resizebox{1.0\textwidth}{!}{
			\begin{tabular}{r|ccc|c|ccc|ccc}\toprule
				\multirow{2}[0]{*}{\textrm{Method}} & \multicolumn{1}{c}{\multirow{2}[0]{*}{\textrm{OF}}} & \multicolumn{1}{c}{\multirow{2}[0]{*}{\textrm{RGB}}} & \multicolumn{1}{c|}{\multirow{2}[0]{*}{\textrm{MF}}} &  $\mathcal{J}\&\mathcal{F}$  & \multicolumn{3}{c|}{$\mathcal{J}$} & \multicolumn{3}{c}{$\mathcal{F}$} \\
				&       &   &    & \textrm{Mean}$\uparrow$  & \textrm{Mean}$\uparrow$  & \textrm{Recall}$\uparrow$ & \textrm{Decay}$\downarrow$ & \textrm{Mean}$\uparrow$  & \textrm{Recall}$\uparrow$ & \textrm{Decay}$\downarrow$ \\ \midrule \midrule
				\textrm{SFL} \cite{SFL}    &       &   \checkmark  & \checkmark & 67.1  & 67.4  & 81.4  & 6.2   & 66.7  & 77.1  & 5.1  \\
				\textrm{LMP} \cite{LMP}    &    \checkmark   &    &   & 68.0  & 70.0  & 85.0  & \textcolor[rgb]{ 0,  0,  1}{1.3}   & 65.9  & 79.2  & 2.5  \\
				\textrm{FSEG} \cite{FSEG}   &   \checkmark    &  \checkmark  &   & 68.0  & 70.7  & 83.0  & 1.5   & 65.3  & 73.8  & 1.8  \\
				\textrm{LVO} \cite{LVO}   &   \checkmark    &   \checkmark  &  & 74.0  & 75.9  & 89.1  & \textcolor[rgb]{ 1,  0,  0}{0.0}   & 72.1  & 83.4  & 1.3  \\
				\textrm{PDB} \cite{PDB}   &       &     \checkmark &  & 75.9  & 77.2  & 90.1  & \textcolor[rgb]{ 0,  1,  0}{0.9}   & 74.5  & 84.4  & \textcolor[rgb]{ 1,  0,  0}{-0.2}  \\
				\textrm{MOT} \cite{MotAdapt}   &   \checkmark    &   \checkmark  &  & 77.3  & 77.2  & 87.8  & 5.0   & 77.4  & 84.4  & 3.3  \\
				\textrm{EPO} \cite{EpONet}   &   \checkmark    &   \checkmark & \checkmark  & 78.1  & 80.6  & \textcolor[rgb]{ 0,  0,  1}{95.2}  & 2.2   & 75.5  & 87.9  & 2.4  \\
				\textrm{FEM} \cite{FEMNet}   &    \checkmark   &    \checkmark &  & 78.4  & 79.9  & 93.9  & 4.4   & 76.9  & 88.3  & 2.4  \\
				\textrm{AGS} \cite{AGS}   &       &   \checkmark  &  & 78.6  & 79.7  & 91.1  & 1.9   & 77.4  & 85.8  & 1.6  \\
				\textrm{AGNN} \cite{AGNN}  &       &    \checkmark & \checkmark & 79.9  & 80.7  & 94.0  & \textcolor[rgb]{ 1,  0,  0}{0.0}   & 79.1  & \textcolor[rgb]{ 0,  0,  1}{90.5}  & \textcolor[rgb]{ 0,  1,  0}{0.0} \\
				\textrm{COS} \cite{COSNet1}   &       &   \checkmark  & \checkmark & 80.0  & 80.5  & 93.1  & 4.4   & 79.4  & 90.4  & 5.0  \\
				\textrm{COS}$^{\dagger}$ \cite{COSNet2}   &       &   \checkmark  & \checkmark & 80.4  & 81.1  & 93.8  & 5.3   & 79.7  & \textcolor[rgb]{ 0,  1,  0}{91.1}  & 5.6  \\
				\textrm{AnDiff}$^{*}$ \cite{ADNet} &       &   \checkmark  & \checkmark & 81.1  & 81.7  & 90.9  & 2.2   & 80.5  & 85.1  & \textcolor[rgb]{ 0,  0,  1}{0.6}  \\
				\textrm{DyStaB} \cite{DyStaB} &    \checkmark   &   \checkmark  &  & 81.4  & 82.8  & -     & -     & 80.0  & -     & - \\
				\textrm{WCS} \cite{WCSNet}   &       &  \checkmark  & \checkmark  & 81.5  & 82.2  & -     & -     & 80.7  & -     & - \\
				\textrm{MAT} \cite{MATNet}   &   \checkmark    &   \checkmark  &  & 81.6  & 82.4  & 94.5  & 5.5   & 80.7  & 90.2  & 4.5  \\
				\textrm{DFNet}$^{*}$\cite{DFNet} &       &   \checkmark  & \checkmark & \textcolor[rgb]{ 0,  0,  1}{82.6}  & 83.4  & -     & -     & \textcolor[rgb]{ 0,  0,  1}{81.8}  & -     & - \\
				\textrm{TransportNet} \cite{TransportNet} &   \checkmark     &   \checkmark  &  & \textcolor[rgb]{ 1,  0,  0}{84.8}  & \textcolor[rgb]{ 0,  1,  0}{84.5}  & -     & -     & \textcolor[rgb]{ 1,  0,  0}{85.0}  & -     & - \\
				\textrm{RTNet} \cite{RTNet} &   \checkmark     &   \checkmark &  \checkmark & \textcolor[rgb]{ 0,  1,  0}{84.2}  & \textcolor[rgb]{ 1,  0,  0}{84.8}  & \textcolor[rgb]{ 0,  1,  0}{95.8}     & -     & \textcolor[rgb]{ 0,  1,  0}{83.5}  & \textcolor[rgb]{ 1,  0,  0}{93.1}     & - \\ \midrule
				\textrm{\textbf{Ours}} &       &   \checkmark  & \checkmark  & 81.9  & 82.7  & 94.8  & 3.6   & 81.1  & 88.6  & 3.6  \\ 
				\textrm{\textbf{Ours}}$^{*}$ &       &   \checkmark   & \checkmark & \textcolor[rgb]{ 0,  0,  1}{82.6}  & \textcolor[rgb]{ 0,  0,  1}{83.5}  & \textcolor[rgb]{ 1,  0,  0}{95.9}  & 4.6   & 81.7  & 89.4  & 4.4  \\ \bottomrule
			\end{tabular}%
		}
		\begin{tablenotes}
			\footnotesize
			\item[*] Uses multi-scale inference and instance pruning, and our method only uses the former.
			%	$^{*}$ Uses multi-scale inference and instance pruning.
			\item[$\dagger$] COS is improved by group attention mechanisms.
		\end{tablenotes}
	\end{threeparttable}
	\label{tab:davis16}%
\end{table*}%

\subsubsection{Evaluation on $\textit{DAVIS}_{\textit{16}}$}
\
\newline
\noindent \textbf{Quantitative result.}
Table \ref{tab:davis16} shows the detailed results, with several top-performing UVOS methods, including single-modality input methods \cite{SFL, LMP, PDB, COSNet1, COSNet2, ADNet, AGNN, AGS, WCSNet, DFNet} and multi-modality input models \cite{LVO, LSMO, FSEG, MotAdapt, MATNet, EpONet, FEMNet, DyStaB, TransportNet, RTNet} taken from the $\textit{DAVIS}_{\textit{16}}$ benchmark. We can observe that our IMCNet achieves competitive performance compared to other methods. As shown in Table \ref{tab:davis16}, our model achieves the third-best results in terms of mean $\mathcal{J}\&\mathcal{F}$ and we can find that IMCNet has achieved the best results $95.9\%$ in terms of recall value of $\mathcal{J}$. Specifically, our IMCNet outperforms all single-modality-based methods and achieves the results on the $\textit{DAVIS}_{\textit{16}}$ with $\mathbf{82.6\%}$ over mean $\mathcal{J}\&\mathcal{F}$, and is equal to the DFNet \cite{DFNet}. In addition, we do not apply CRF post-processing. The results indicate that our model can capture motion information and implicitly compensate motion by the MCM better than the single-modality input method. Multi-modality methods \cite{TransportNet, RTNet} use the optical flow as a cue to segment objects and can better capture the motion information. Therefore, these two multi-modality methods, \emph{i.e.}, TransportNet\cite{TransportNet} and RTNet \cite{RTNet}, achieved the best results in the UVOS task. Although the multi-modality input methods TransportNet\cite{TransportNet} and RTNet \cite{RTNet} have achieved the best result, due to introducing an additional pre-processing stage to predict the optical flow, the number of model parameters is more than our model, and inference time is slower (See \S \ref{subsubsec:run}).

\begin{figure*}[htb]
	\begin{center}
		\includegraphics[width=0.95\linewidth]{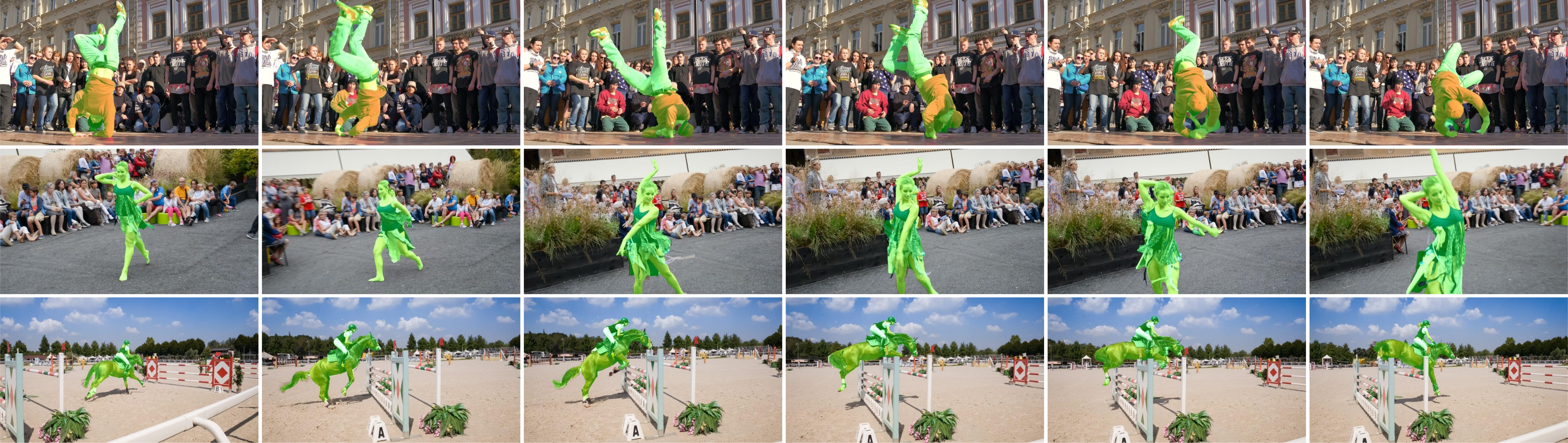}
	\end{center}
	\caption{Qualitative results on three videos from the $\textit{DAVIS}_{\textit{16}}$ dataset. From top to bottom: \textit{breakdance}, \textit{dance-twirl}, and \textit{horsejump-high}.}
	\label{fig:davis_qual}
\end{figure*}

\noindent
\textbf{Qualitative results.}
Fig. \ref{fig:davis_qual} depicts the qualitative results on $\textit{DAVIS}_{\textit{16}}$ which contains some challenges like cluttered background, deformation, and motion blur, \emph{e.g.}, \textit{breakdance}, \textit{dance-twirl}, and \textit{horsejump-high}. As seen, our IMCNet is robust to these challenges and precisely extracts primary objects(s) with accurate boundaries. The \textit{breakdance} and \textit{dance-twirl} sequences from the $\textit{DAVIS}_{\textit{16}}$ contain similar surroundings and large deformation. We can find that our method can effectively discriminate the target from those background distractors, thanks to the ACM and APM. Besides, through implicit learned and compensating motion by the MCM, our IMCNet can accurately segment some sequences (\emph{e.g.}, \textit{horsejump-high}), where there is fast motion and suffer from motion blur.

\subsubsection{Evaluation on $\textit{YouTube-Objects}$}
\
\newline
\noindent \textbf{Quantitative result.}
Table \ref{tab:youtubeobj} reports the results of several compared methods \cite{LVO, SFL, FSEG, PDB, AGS, COSNet1, COSNet2, AGNN, MATNet, WCSNet} for different categories on the $\textit{YouTube-Objects}$ dataset. Our method achieves promising performance in most categories and the second-best overall results on mean $\mathcal{J}$. It is slightly worse than COS$^{\dagger}$ ($\mathbf{-0.5\%}$) in terms of mean $\mathcal{J}$ and achieves the second-best results. However, we emphasize that the COS$^{\dagger}$ is more computationally expensive. Compared with COS, our IMCNet is superior to the COS without group attention mechanisms. It is indicated that COS obtains a precise segmentation mask by capturing richer structure information of videos, but our IMCNet can achieve sufficient segmentation accuracy through three frames without damaging inference speed.

\begin{table*}[!htbp]
	\centering
	\tiny
	\caption{Quantitative performance of each category on the \textit{Youtube-Objects} with the region similarity (mean $\mathcal{J}$). We show the average result for each of the 10 categories from the \textit{Youtube-Objects} and the final row shows an average over all categories. The top three final results are marked in \textcolor[rgb]{1, 0, 0}{Red}, \textcolor[rgb]{0, 1, 0}{Green}, and \textcolor[rgb]{0, 0, 1}{Blue}.}
	\begin{threeparttable}
	\resizebox{1.0\textwidth}{!}{
		\begin{tabular}{r|cccccccccc|cc}\toprule
			Category & LVO \cite{LVO}   & SFL \cite{SFL}  & FSEG \cite{FSEG} & PDB \cite{PDB}  & AGS \cite{AGS}  & COS \cite{COSNet1} & COS$^{\dagger}$ \cite{COSNet2} & AGNN \cite{AGNN} & MAT \cite{MATNet}  & WCS  \cite{WCSNet} & \textbf{Ours} & \textbf{Ours}$^{*}$ \\ \midrule \midrule
			Airplane (6) & 86.2  & 65.6  & 81.7  & 78.0  & 87.7  & 81.1 & 83.4  & 81.1  & 72.9  & 81.8  & 81.2 & 81.1 \\
			Bird (6) & 81.0  & 65.4  & 63.8  & 80.0  & 76.7  & 75.7 & 78.1 & 75.9  & 77.5  & 81.2  & 78.3 & 81.1  \\
			Boat (15) & 68.5  & 59.9  & 72.3  & 58.9  & 72.2  & 71.3 & 71.0 & 70.7  & 66.9  & 67.6  & 67.9 & 70.3 \\
			Car (7) & 69.3  & 64.0  & 74.9  & 76.5  & 78.6  & 77.6 & 79.1  & 78.1  & 79.0  & 79.5  & 78.5 & 77.1  \\
			Cat (16) & 58.8  & 58.9  & 68.4  & 63.0  & 69.2  & 66.5  & 66.1 & 67.9  & 73.7  & 65.8  & 70.4 & 73.3  \\
			Cow (20) & 68.5  & 51.2  & 68.0  & 64.1  & 64.6  & 69.8 & 69.5 & 69.7  & 67.4  & 66.2  & 67.1 & 66.8  \\
			Dog (27) & 61.7  & 54.1  & 69.4  & 70.1  & 73.3  & 76.8  & 76.2 & 77.4  & 75.9  & 73.4  & 73.1 & 74.8  \\
			Horse (14) & 53.9  & 64.8  & 60.4  & 67.6  & 64.4  & 67.4  & 71.9 & 67.3  & 63.2  & 69.5  & 63.0 & 64.8  \\
			Motorbile (10) & 60.8  & 52.6  & 62.7  & 58.3  & 62.1  & 67.7 & 67.9 & 68.3  & 62.6  & 69.3  & 63.3 & 58.7  \\
			Train (5) & 66.3  & 34.0  & 62.2  & 35.2  & 48.2  & 46.8 & 46.5  & 47.8  & 51.0  & 49.7  & 56.8 & 56.8  \\ \midrule
			Mean $\mathcal{J}\uparrow$ & 67.5  & 57.1  & 68.4  & 65.2  & 69.7  & 70.1 & \textcolor[rgb]{ 1,  0,  0}{71.0}  & \textcolor[rgb]{ 0,  0,  1}{70.4}  & 69.0  & \textcolor[rgb]{ 0,  0,  1}{70.4}  & 70.0 & \textcolor[rgb]{ 0,  1,  0}{70.5}  \\ \bottomrule
		\end{tabular}%
	}
\begin{tablenotes}
	\footnotesize
%	\tiny
	\item[*] Uses multi-scale inference.
	%	$^{*}$ Uses multi-scale inference and instance pruning.
	\item[$\dagger$] COS is improved by group attention mechanisms.
\end{tablenotes}
\end{threeparttable}
	\label{tab:youtubeobj}%
\end{table*}%

\begin{figure*}[htb]
	\begin{center}
		\includegraphics[width=0.95\linewidth]{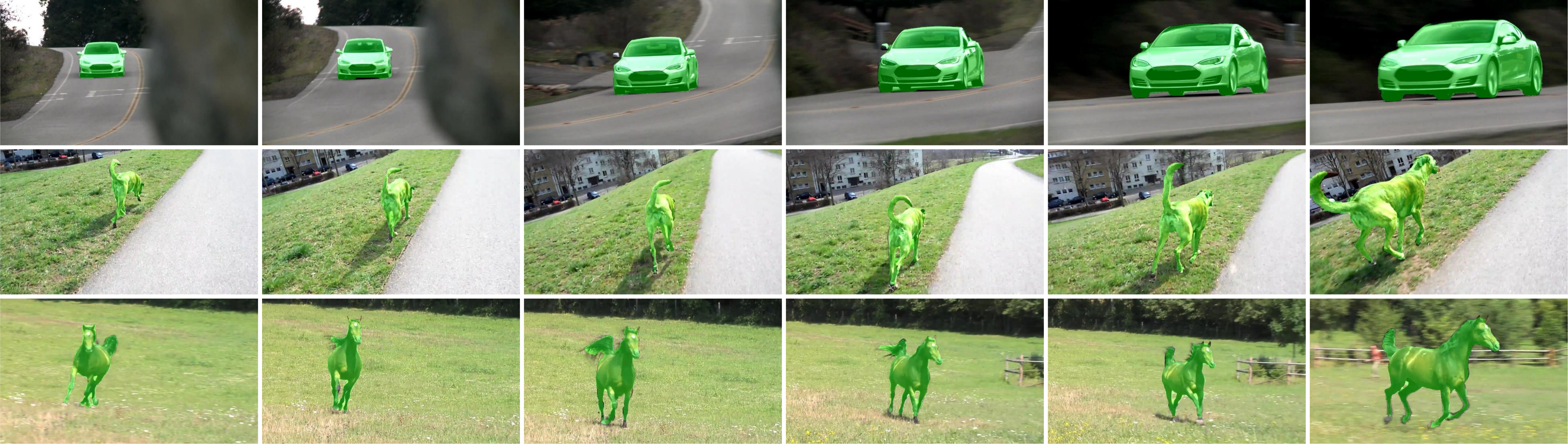}
	\end{center}
	\caption{Qualitative results on three videos from the $\textit{YouTube-Objects}$ dataset. From top to bottom: \textit{car-0009}, \textit{dog-0022}, and \textit{horse-0011}.}
	\label{fig:ytboj_qual}
\end{figure*}

\noindent
\textbf{Qualitative results.}
Fig. \ref{fig:ytboj_qual} shows the qualitative results on $\textit{YouTube-Objects}$. We can observe that the target object suffering some challenging scenarios like fast motion (\emph{e.g.}, \textit{car-0009} and \textit{horse-0011}) and large deformation (\emph{e.g.}, \textit{dog-0022} and \textit{horse-0011}). Our model can deal with such challenges well, it verifies the effectiveness of the MCM.

\subsubsection{Runtime comparison}\label{subsubsec:run}

To further investigate the computation efficiency of our IMCNet, we report the number of network parameters and inference time comparisons on the $\textit{DAVIS}_{\textit{16}}$ datasets with a 480p resolution. We do not count data loading and only focus on the segmentation time of the models. We compare the IMCNet with state-of-the-art methods which share their codes or include the corresponding experimental results, including AGNN \cite{AGNN}, AnDiff \cite{ADNet}, COS \cite{COSNet1}, MAT \cite{MATNet} DFNet \cite{DFNet}, TransportNet \cite{TransportNet}, and RTNet \cite{RTNet}. For the inference time comparison, we run the public code of other methods and our code on the same conditions with NVIDIA TITAN RTX GPU. The analysis results are summarized in Table \ref{tab:runcom}.

\begin{figure}[htbp]
	\begin{center}
		\includegraphics[width=0.8\linewidth]{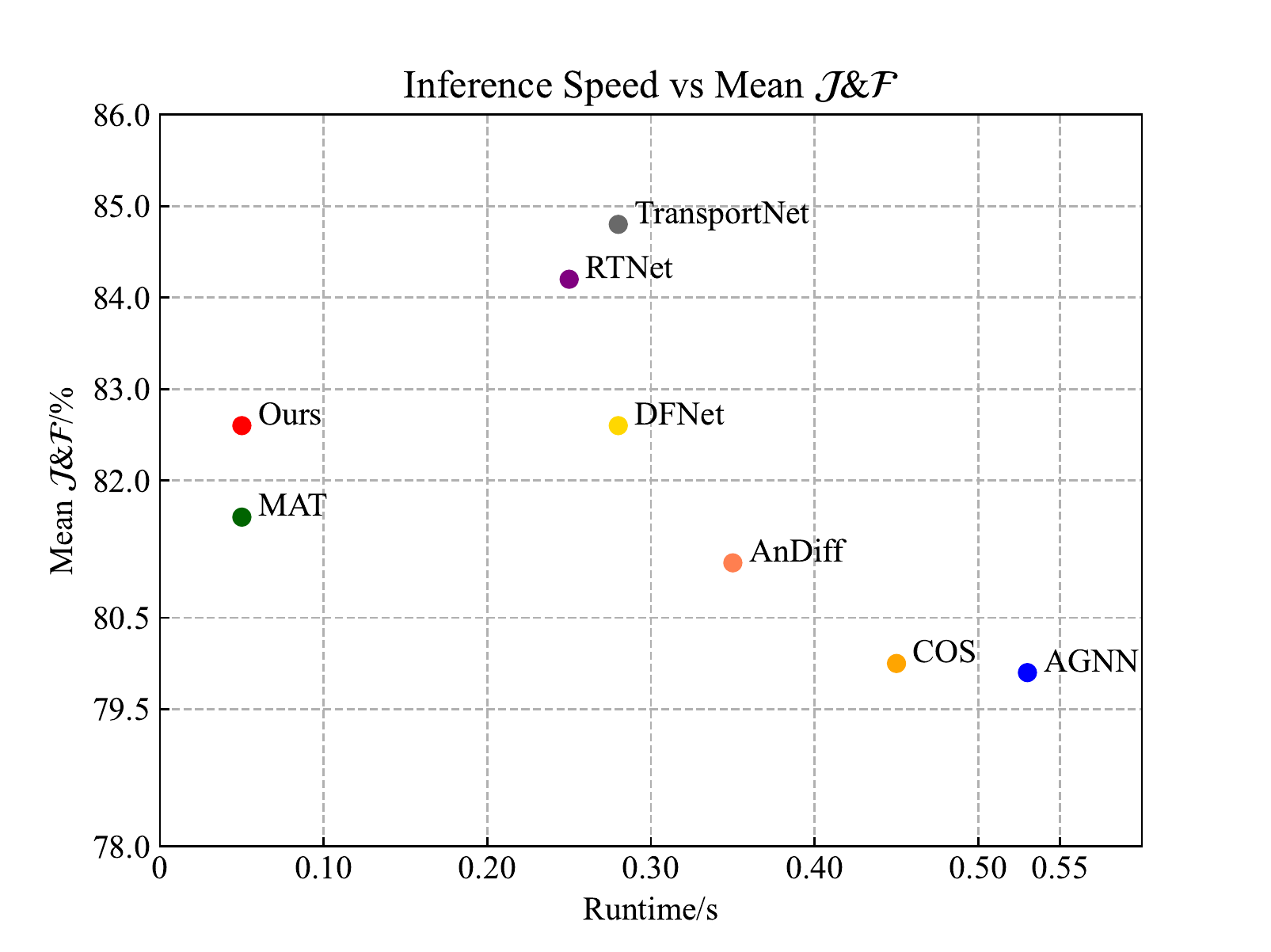}
	\end{center}
	\caption{Trade-off between inference time (x-axis) and segmentation accuracy (y-axis) on $\textit{DAVIS}_{\textit{16}}$. Our approach demonstrates compelling performance with high efficiency}
	\label{fig:speed}
\end{figure}

Table \ref{tab:runcom} shows that our IMCNet reduces the model complexity with fewer parameters than the other methods. For the inference comparison, we can observe that our method shows a faster speed than other competitors. Fig. \ref{fig:speed} depicts a visualization of the trade-off between accuracy and efficiency of representative methods on the validation set of  $\textit{DAVIS}_{\textit{16}}$. As can be seen, our IMCNet achieves the best trade-off.

\begin{table}[htb]
	\centering
	\scriptsize
	\caption{The number of model parameters and inference time comparison with state-of-the-art methods. The abbreviation `M' in the `\#Param.' cell represents a million.}
	\resizebox{1.0\columnwidth}{!}{
		\begin{threeparttable}
		\begin{tabular}{c|ccccccc|c} \toprule
			Method & AGNN \cite{AGNN}  & AnDiff \cite{ADNet} & COS \cite{COSNet1}  & MAT \cite{MATNet}   & DFNet \cite{DFNet} & TransportNet \cite{TransportNet}& RTNet\tnote{1} \cite{RTNet}& \textbf{Ours} \\ \midrule \midrule
			\#Param. (M) & 82.3  & 79.3  & 81.2  & 142.7 & 64.7 & - & 277.2 & \textbf{47.9} \\ \midrule
			Inf. Time (s/frame) & 0.53  & 0.35  & 0.45  & \textbf{0.05}  & 0.28 & 0.28 & 0.25 & \textbf{0.05} \\ \midrule
			$\textit{DAVIS}_{\textit{16}}$ Mean $\mathcal{J}\&\mathcal{F}\uparrow$& 79.9  & 81.1  & 80.0  & 81.6  & 82.6 & \textbf{84.8} & 84.2 & 82.6 \\ \bottomrule
		\end{tabular}%
	\begin{tablenotes}
		\footnotesize
		\item[1] The RTNet source code only provides a pre-trained model based on ResNet-34, we use the model based on ResNet-34 here.
	\end{tablenotes}
	\end{threeparttable}
	}	
	\label{tab:runcom}%
\end{table}%

\section{Conclusion}\label{sec:con}

In this paper, we proposed a novel framework, IMCNet, for the UVOS. The proposed IMCNet mines the long-term correlations from several input frames by a light-weight affinity computing module. In addition, an attention propagation module is proposed to transmit global correlation in a top-down manner. Finally, a novel motion compensation module aligns motion information from temporally adjacent frames to the current frame which achieves implicit motion compensation at the feature level. Experimental results demonstrated that the proposed IMCNet achieves favorable performance against other methods while running at a faster speed and using much fewer parameters.

\section*{Acknowledgments}
This work was supported by the National Nature Science Foundation of China under grant 61620106012 and Beijing Natural Science Foundation under grant 4202042.

\bibliographystyle{IEEEtran}
\bibliography{mybib}

\end{document}